\documentclass[lettersize,journal]{IEEEtran}
\usepackage[switch]{lineno}
\usepackage{amsmath,amsfonts}
\usepackage{algorithmic}
\usepackage{algorithm}
\usepackage{array}
\usepackage[caption=false,font=normalsize,labelfont=sf,textfont=sf]{subfig}
\usepackage{textcomp}
\usepackage{stfloats}
\usepackage{url}
\usepackage{verbatim}
\usepackage{graphicx}
\usepackage{cite}
\usepackage{multirow}
\usepackage{booktabs}
\usepackage{amsbsy}
\usepackage{rotating}
\usepackage{bigstrut}
\usepackage{color}
\usepackage[table]{xcolor}
\usepackage{tikz,orcidlink}
\hypersetup{
	colorlinks=true,
	linkcolor=black,
	citecolor=black
}
\let\oldequation\equation
\let\oldendequation\endequation
\renewenvironment{equation}{\linenomathNonumbers\oldequation}{\oldendequation\endlinenomath}

\begin{document}

\title{Alignment-Free RGBT Salient Object Detection: Semantics-guided Asymmetric Correlation Network and A Unified Benchmark}

\author{Kunpeng Wang\hspace{-1.5mm}$^{~\orcidlink{0000-0002-2788-7583}}$, Danying Lin\hspace{-1.5mm}$^{~\orcidlink{0009-0001-9993-0448}}$, Chenglong Li\hspace{-1.5mm}$^{~\orcidlink{0000-0002-7233-2739}}$, Zhengzheng Tu\hspace{-1.5mm}$^{~\orcidlink{0000-0002-9689-8657}}$, Bin Luo\hspace{-1.5mm}$^{~\orcidlink{0000-0001-5948-5055}}$,~\IEEEmembership{Senior~Member,~IEEE} 
\thanks{This work is supported in part by University Synergy Innovation Program of Anhui Province (No.GXXT-2022-014), in part by National Natural Science Foundation of China under Grant 62376005, in part by National Natural Science Foundation of China (No. 62376004), in part by Natural Science Foundation of Anhui Province (No. 2208085J18), in part by the Natural Science Foundation of Anhui Higher Education Institution of China under Grant KJ2020A0033, in part by Anhui Provincial Natural Science foundation under Grant 2108085MF211, in part by Anhui Energy Internet Joint Fund Project under Grant 2008085UD07, in part by the National Natural Science Foundation of China under Grant 61876002, in part by Anhui Provincial Key Research and Development Program under Grant 202104d07020008, and in part by the NSFC Key Project of International (Regional) Cooperation and Exchanges under Grant 61860206004. (Corresponding author is Zhengzheng Tu)}
\thanks{Kunpeng Wang, Danying Lin, Zhengzheng Tu and Bin Luo are affiliated with Information Materials and Intelligent Sensing Laboratory of Anhui Province, Anhui Provincial Key Laboratory of Multimodal Cognitive Computation, School of Computer Science and Technology, Anhui University, Hefei 230601, China (e-mail: kp.wang@foxmail.com; danying\_lin@foxmail.com; zhengzhengahu@163.com and luobin@ahu.edu.cn).}
\thanks{Chenglong Li is affiliated with the Information Materials and Intelligent  Sensing Laboratory of Anhui Province, Anhui Provincial Key Laboratory  of Security Artificial Intelligence, School of Artificial Intelligence, Anhui  University, Hefei 230601, China (e-mail: lcl1314@foxmail.com).}}

\markboth{Journal of \LaTeX\ Class Files,~Vol.~14, No.~8, August~2021}%
{Shell \MakeLowercase{\textit{et al.}}: A Sample Article Using IEEEtran.cls for IEEE Journals}


\maketitle

\begin{abstract}
RGB and Thermal (RGBT) Salient Object Detection (SOD) aims to achieve high-quality saliency prediction by exploiting the complementary information of visible and thermal image pairs, which are initially captured in an unaligned manner. However, existing methods are tailored for manually aligned image pairs, which are labor-intensive, and directly applying these methods to original unaligned image pairs could significantly degrade their performance. In this paper, we make the first attempt to address RGBT SOD for initially captured RGB and thermal image pairs without manual alignment. Specifically, we propose a Semantics-guided Asymmetric Correlation Network (SACNet) that consists of two novel components: 1) an asymmetric correlation module utilizing semantics-guided attention to model cross-modal correlations specific to unaligned salient regions; 2) an associated feature sampling module to sample relevant thermal features according to the corresponding RGB features for multi-modal feature integration. In addition, we construct a unified benchmark dataset called UVT2000, containing 2000 RGB and thermal image pairs directly captured from various real-world scenes without any alignment, to facilitate research on alignment-free RGBT SOD. Extensive experiments on both aligned and unaligned datasets demonstrate the effectiveness and superior performance of our method. The dataset and code are available at \href{https://github.com/Angknpng/SACNet}{https://github.com/Angknpng/SACNet}.
\end{abstract}


\section{Introduction}
Salient Object Detection (SOD) aims to identify and segment the most attractive regions in visible images. It has been applied in a variety of tasks, such as image retrieval~\cite{ling2006retrieval}, person re-identification~\cite{zhao2013reid}, object tracking~\cite{zhang2020track}, and video analysis~\cite{kong2021self,fan2019video}. Despite the great success achieved by existing methods~\cite{yun2023towards,zhuge2022salient,piao2022noise}, single-modal SOD remains challenging when dealing with deteriorated visible images caused by illumination variation~\cite{zhu2023cross,xu2021exploring}, complex background, etc. Since thermal images are captured based on the infrared radiation energy of the scene, they are not interfered by illumination and can capture the overall shape of objects. RGB and Thermal (RGBT) SOD improves the performance of single-modal SOD by introducing the corresponding thermal modality and exploiting their complementary benefits. 
\begin{figure}[t]
	\centering
	\includegraphics[width=1\columnwidth]{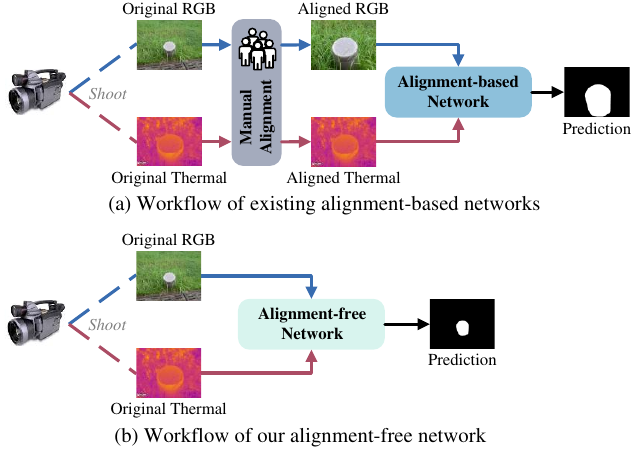}
	\caption{Workflow comparisons between existing networks and our network. (a) Existing networks require a labor-intensive manual alignment process to align visible and thermal image pairs, further exploiting modality complementarity for saliency prediction. (b) Our network directly mines multi-modal correlations and complementarity of initially captured unaligned image pairs for saliency prediction.} 
	\label{fig::motivation}
\end{figure}
In practice, the RGBT image pairs captured directly by the device are unaligned. However, existing RGBT SOD datasets are manually aligned, which consumes a lot of labor and fails to reflect the problems caused by unalignment in practical applications. Based on this, existing methods are almost tailored for aligned RGBT image pairs and are difficult to adapt to unaligned cases, which directly leads to performance degradation on unaligned image pairs.

Specifically, existing RGBT SOD methods~\cite{wan2023mffnet,zhang2019rgb,tu2021multi,wang2021cgfnet,liu2021swinnet,zhou2023wave} integrate the complementary information of the two modalities through different fusion paradigms. Although promising results are achieved, they are designed upon manually aligned visible and thermal image pairs, which requires heavy labor costs. Fig.~\ref{fig::motivation} (a) illustrates the workflow of existing methods that perform saliency prediction on the basis of manually aligned RGBT image pairs. In the case of alignment, objects in RGB and thermal modalities inherently correspond in space, they are naturally correlated, and simple operations such as summation and concatenation can directly exploit the multi-modal complementarity. However, the initially captured image pairs in practice are unaligned~\cite{tu2019rgb}, exhibiting deviations in both position and scale, as shown in Fig.~\ref{fig::camera}. In such case, the correlation between the two modalities significantly decreases, making it challenging to directly exploit multi-modal complementarity. Therefore, applying existing methods to initially captured image pairs without alignment can severely degrade their performance.

Recently, DCNet~\cite{tu2022weakly} attempts to address this issue for weakly aligned image pairs. It performs random spatial affine transformations on existing aligned datasets~\cite{wang2018rgb,tu2019rgb,tu2020rgbt} to artificially create weakly aligned datasets. Based on this, it models the correlation between weakly aligned image pairs through dynamic convolution and feature-wise affine transformation. Nonetheless, two issues still exist: 1) affine transformation with limited transformation space and dynamic convolution with small receptive field are difficult to deal with large spatial deviations, which exist in the initially captured image pairs; 2) artificially created weakly aligned datasets have small deviations and lack some practical significance. Consequently, DCNet also fails when applied to the directly captured unaligned image pairs.
For the first issue, we model the comprehensive correlation between the two modalities through an asymmetric-window based correlation operation, which fully associates the corresponding unaligned multi-modal information. Based on this, we sample and integrate relevant multi-modal features through cascaded deformable convolutions for accurate saliency prediction. 
For the second issue, we construct an unaligned dataset through real-world camera shooting to facilitate the research of alignment-free RGBT SOD.

To this end, we propose a Semantics-guided Asymmetric Correlation Network (SACNet) to enhance the performance of alignment-free RGBT SOD. 
Specifically, we propose an Asymmetric Correlation Module (ACM) based on transformer attention~\cite{vaswani2017attention} to model comprehensive multi-modal correlations.
since the two modalities are unaligned in position and scale, the ACM restricts the correlation modeling within a pair of asymmetric windows, which preserves complete corresponding salient regions of the two modalities, thereby reducing the inconsistency interference caused by misalignment. To further reduce the interference of background noise, semantic information is embedded into the ACM to guide the correlation modeling to focus on salient regions. In addition, the AFSM samples relevant thermal features conditioned on RGB features through cascaded deformable convolutions, enabling the integration for corresponding multi-modal information. Guided by the above two modules, our method can model robust correlations between the two modalities to further exploit the multi-modal complementary benefits for accurate saliency prediction. As shown in Fig.~\ref{fig::motivation} (b), our method is able to predict the saliency maps for directly captured unaligned image pairs.

By designing the ACM and AFSM modules, we propose the correlation modeling technique that associates and samples the corresponding information in unaligned RGBT image pairs, which are captured directly in the real world without manual alignment. There are several good impacts and applications for the proposed technique. First, based on this technique, our method achieves alignment-free RGBT SOD, which saves the labor costs caused by manual alignment. Second, the proposed correlation modeling technique can be extended to other multi-modal tasks to enhance the correlation between modalities and improve their performance, such as RGBT tracking~\cite{liu2023non}, RGBT super-resolution~\cite{zhao2023thermal}, and image-text retrieval~\cite{jiang2023cross}. Third, the proposed technique provides a basis for the collaborative utilization of multi-modal information. Since manual alignment is not required, the proposed technique can be combined with other techniques and deployed into devices with RGB and thermal sensors (e.g., surveillance cameras, satellites, and UAVs) to achieve intelligent video surveillance, remote sensing image analysis, UAV monitoring, etc.

In addition, we construct a unified benchmark dataset with practical significance, UVT2000, to facilitate research on alignment-free RGBT SOD. UVT2000 contains 2000 unaligned visible-thermal image pairs with ground truth annotations, directly captured by a pair of thermal infrared and CCD cameras without any alignment. Therefore, the misalignment of the image pairs is a natural result of camera shooting and reflects the issues in practical applications. Additionally, the image pairs are collected from a variety of real-world scenarios, which are annotated with 11 challenge attributes to facilitate the study of specific issues.

To the best of our knowledge, this is the first work to launch the alignment-free setting and the corresponding benchmark dataset for RGBT SOD. The main contributions of our work are as follows:
\begin{itemize}
	\item For the first time, we perform RGBT SOD on initially captured visible-thermal image pairs without any manual alignment, which can significantly reduce labor costs.
	\item We propose an Asymmetric Correlation Module (ACM) to model multi-modal correlations specific to salient regions, and an Associated Feature Sampling Module (AFSM) to sample and integrate relevant features of the two modalities.
	\item We construct a novel benchmark dataset, containing 2000 unaligned visible-thermal image pairs directly captured from various real-word scenes, to facilitate research on alignment-free RGBT SOD.
	\item Our proposed method achieves state-of-the-art performance on both aligned and unaligned datasets, demonstrating its effectiveness and the potential of alignment-free RGBT SOD.
\end{itemize}

\section{Related Work}
\subsection{RGBD Salient Object Detection}
In the past decades, a large number of Salient Object Detection (SOD) methods~\cite{wang2022survey} have been developed through feature refinement~\cite{Wu19CPD,liu2022poolnet+,xiao2018deep}, attention mechanism~\cite{liu2020picanet}, boundary enhancement~\cite{qin2019basnet,yao2021boundary}, uncertainty perception~\cite{tian2023modeling,wang2022multi}, etc. However, they still struggle to handle some challenging scenes such as similar foreground and background, low illumination, and image clutter. To address these issues, some studies~\cite{peng2014benchmark,qu2017deep} initially introduce depth maps with spatial structure information to enhance the performance of single-modal SOD, called RGBD SOD. In order to exploit the multi-modal complementary information, existing methods~\cite{zhou21survey} mainly fuse visible images and depth maps through early fusion, middle fusion, and late fusion.

Early fusion integrates visible images and depth maps into a joint representation as input to a network. For example, Qu et al.~\cite{qu2017deep} compute the joint prior features of visible images and depth maps as input to a convolutional neural network to extract a unified multi-modal feature representation. Song et al.~\cite{song2017daware} predict multi-level saliency maps based on multi-scale pre-segmentation results of the input RGBD image. 
Middle fusion mainly utilizes multi-scale feature fusion strategies to mine multi-modal correlations. BBSNet~\cite{fan2020bbsnet} divides multi-level features into teacher and student features, and utilizes the discriminative semantics of the teacher features to suppress the interference in the student features. Cong et al.~\cite{cong2022cirnet} effectively utilize multi-modal information by progressively integrating multi-level features in the encoder and decoder based on the attention mechanism. Wen et al.~\cite{wen2023cross} improve the universality and anti-interference of saliency predictions by enhancing extracted features and inferring high-level semantic information.
Late fusion learns high-level features or saliency maps of the two modalities for fusion. Han et al.~\cite{han2018late} learn the feature representations of RGB and depth modalities separately, and mine their complementary relationships to obtain a joint representation for saliency prediction. Zhang et al.~\cite{zhang2022c} independently enhance intra-modal features of two modalities and then selectively interact them based on scene information.

Although these methods exploit the complementary information of the RGB and depth modalities through different fusion strategies, they are designed on two  well-aligned modalities, making it difficult to transfer to unaligned multi-modal inputs for correlation modeling.

\subsection{RGBT Salient Object Detection}
Since thermal images are captured based on the infrared radiation of objects without suffering from the interference of complex environments (e.g., illumination and fog), several recent researches~\cite{wang2018rgb,tu2019rgb,tu2020rgbt} introduce the thermal modality based on the RGB modality to form RGBT SOD.
Existing methods mainly focus on mining modality complementarity~\cite{tu2021multi,wang2021cgfnet,huang2021multi,liu2021swinnet}, alleviating the modality gap~\cite{chen2022cgmdrnet,xie2023cross}, or addressing modality-specific challenges such as low illumination~\cite{zhang2019rgb,zhou2021ecffnet,zhou2022apnet} and thermal crossover~\cite{liao2022cross}, etc.

For example, to handle the challenge of low illumination, Zhang et al.~\cite{zhang2019rgb} integrate multi-level and multi-modal features, Liao et al.~\cite{liao2022cross} facilitate the multi-modal interaction in the encoder for discriminative feature extraction. To exploit modality complementarity for accurate prediction, Tu et al.~\cite{tu2021multi} design a dual-decoder and a multi-type interaction. Wang et al~\cite{wang2021cgfnet} excavate multi-modal complementary information with unique single-modal information for mutual guidance. Liu et al.~\cite{liu2021swinnet} utilize the strong feature representation capability of Swin Transformer to alleviate the gap between the two modalities. For practical applications, Zhou et al.~\cite{zhou2023lsnet} begin to reduce computational complexity with lightweight operations. Although these methods have achieved success in addressing different issues, they all perform on manually aligned RGBT image pairs, incurring expensive labor annotation costs. 

Recently, Tu et al.~\cite{tu2022weakly} initially address this issue for artificially weakly aligned image pairs by employing affine transformation and dynamic convolution. Although shown to be effective, the limited transform space of affine transformation and the small receptive field of dynamic convolution limit the correlation modeling for large spacial deviations. Additionally, the artificially created datasets lack practical significance. Different from existing models, we directly address RGBT SOD for original captured visible and thermal image pairs without any manual alignment, and construct a novel unaligned dataset to validate the effectiveness of our method.

\subsection{Attention-based Methods}
Since attention mechanism~\cite{vaswani2017attention,Dosovitskiy21vit} has powerful global context modeling capabilities, they have been applied to various fields, such as visual tracking~\cite{chen2021transt}, object detection~\cite{Carion2020detr}, semantic segmentation~\cite{zheng21setr}, and image super-resolution~\cite{zhou2023srformer}, demonstrating competitive performance. For example, Carion et al.~\cite{Carion2020detr} regard the object detection task as a set prediction problem and build a transformer encoder-decoder framework for detection that eliminates unnecessary manually designed components in the detection process. Zheng et al.~\cite{zheng21setr} apply the attention mechanism to the segmentation field pioneeringly. They use a vision transformer~\cite{Dosovitskiy21vit} to extract hierarchical features, which are fed into a decoder to obtain predictions. Chen et al.~\cite{chen2021transt} design a transformer tracker that uses transformer attention to establish associations between template and search region features to highlight useful target information.

Some recent methods~\cite{liu2021visual,liu2021swinnet,pang2023caver,Tang2023RTransNet} have also shown the effectiveness of attention mechanism in SOD tasks. 
For example, Liu et al.~\cite{liu2021visual} construct the first model based on a pure transformer architecture for RGB and RGBD SOD tasks. By introducing task-related tokens and patch-task-attention in the decoder, boundary and saliency prediction are jointly implemented. Tang et al.~\cite{Tang2023RTransNet} fuse the information of two modalities at the input level and send it into a high-resolution transformer, which can maintain high resolution and preserve large receptive fields, to interact multi-modal features for saliency prediction. Pang et al.~\cite{pang2023caver} improve the computation efficiency of transformer attention by aggregating and converting pixel-level tokens to patch-level tokens before the multiplication operation of attention. Based on it, an efficient top-down transformer-based information propagation strategy is designed to integrate multi-modal features. 

Nevertheless, to the best of our knowledge, there is no attention-based method designed to model the correlation of salient objects in unaligned image pairs. In this paper, we propose a semantics-guided asymmetric attention to model unaligned multi-modal correlations. 

\begin{figure*}[t]
	\centering
	\includegraphics[width=1\linewidth]{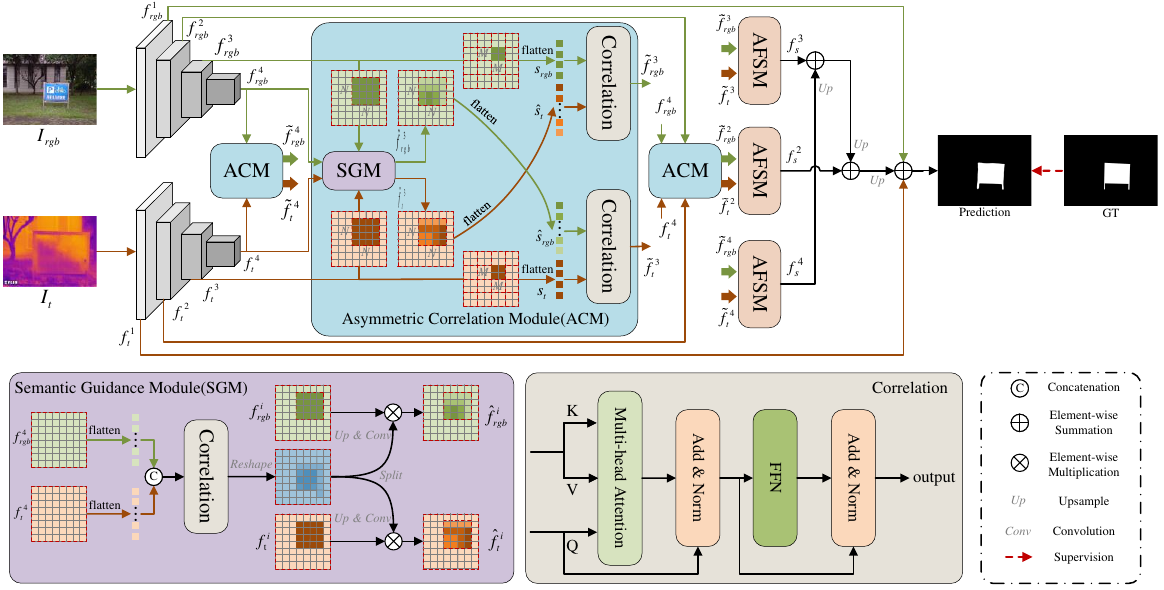}
	\caption{The overall architecture of our proposed SACNet. The framework mainly comprises an Asymmetric Correlation Module (ACM) and an Associated Feature Sampling Module (AFSM). The ACM restricts the correlation operation within asymmetric window pairs to model comprehensive correlations of the two unaligned modalities. With the Semantic Guidance Module (SGM), ACM focus more on salient regions. In the AFSM, relevant thermal features are sampled according to corresponding RGB features. Subsequently, the multi-modal saliency cues are integrated complementarily for saliency prediction.} 
	\label{fig::framework}
\end{figure*}
\begin{figure}[t]
	\centering
	\includegraphics[width=1\columnwidth]{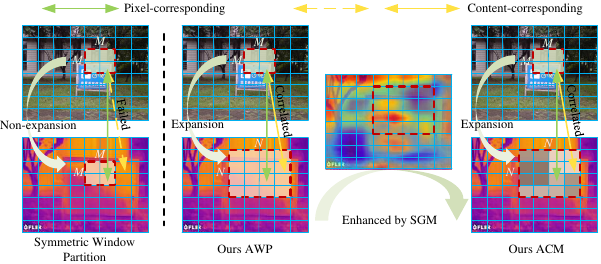}
	\caption{Comparison of correlation modeling in unaligned RGBT image pairs between the conventional symmetric window partition and the proposed Asymmetric Correlation Module (ACM), which contains the Asymmetric Window Partition (AWP) strategy and Semantic Guidance Module (SGM).} 
	\label{fig::sample}
\end{figure}
\section{Our Method: SACNet}
In this paper, we propose a Semantics-guided Asymmetric Correlation Network (SACNet) to model strong multi-modal correlations, thereby leveraging the complementary information of unaligned image pairs for saliency prediction. In the following, we present an overview of SCANet in Section~\ref{Overview}, describe the details of ACM and AFSM in Sections~\ref{ACM} and~\ref{AFSM}, respectively, and formulate the loss function in Section~\ref{loss}.

\subsection{Overview} 
\label{Overview}
In this section, we describe the pipeline of the proposed method for alignment-free RGB and Thermal (RGBT) salient object detection (SOD). The overall architecture of our framework is shown in Fig.~\ref{fig::framework}, which consists of two parallel backbones for multi-modal feature extraction, an Asymmetric Correlation Module (ACM), and an Associated Feature Sampling Module (AFSM). The multi-level features extracted by the backbones are collected and denoted as $f_{m}^i$ ($m \in \left\{ {rgb,t} \right\},i = 1,...,4$). As the salient objects in unaligned image pairs have inconsistent positions and scales, the ACM exploits asymmetric window attention with semantic guidance to model comprehensive correlations between the modalities. Then, the AFSM based on cascaded deformable convolutions is proposed to sample and integrate relevant multi-modal features.

\subsection{Asymmetric Correlation Module (ACM)}
\label{ACM}
Existing RGBT Salient Object Detection methods~\cite{tu2021multi,wang2021cgfnet,xie2023cross,zhou2023wave} commonly design different fusion schemes to explore the multi-modal complementarity between aligned image pairs for accurate prediction. Salient regions in aligned image pairs are consistent in spatial location and scale, with strong modality correlation, which facilitates further multi-modal feature integration. Nevertheless, the manual alignment of initially captured unaligned image pairs incurs substantial labor requirements. Furthermore, directly applying these methods to the unaligned image pairs may result in performance degradation, as they struggle to effectively model multi-modal correlations without prior alignment. 

In order to establish robust multi-modal correlations between unaligned image pairs, we propose the Asymmetric Correlation Module (ACM), as shown in Fig.~\ref{fig::framework}. 
Due to the different object positions and scales between the two unaligned modalities, the multi-modal information does not correspond spatially. Thus, dividing the feature maps of the two unaligned modalities with symmetric windows fails to cover complete correspondence information, resulting in insufficient correlation modeling. Moreover, directly modeling correlations over entire feature maps introduces too much undesired background noise. To this end, the ACM divides the feature map of one modality into small windows, and expands each of these windows into a large one on the feature map of another modality to cover the complete corresponding information. In addition, semantic information is introduced to guide the correlation operation within each asymmetric window pair to focus on object information.
In specific, the ACM contains Asymmetric Window Partition (AWP) to alleviate the interference of spatial inconsistency on correlation modeling, and Semantic Guidance Module (SGM) to focus the correlation modeling on salient regions.
Fig.~\ref{fig::sample} vividly demonstrates that ACM can establish sufficient multi-modal correlation for unaligned RGBT image pairs compared to the symmetric window partition.

\subsubsection{Asymmetric Window Partition (AWP)}
As shown in Fig.~\ref{fig::camera}, the salient regions in unaligned image pairs have different scales, resulting in different background contents and proportions. In this case, directly applying the original correlation operation across the entire multi-modal regions might introduce excessive undesired background noise, thereby confusing the correlation modeling for salient regions. To mitigate this concern, we first restrict the correlation operation within a window pair. Then, considering the unknown position shifts and scale differences, we adjust the window pair shape to be asymmetric to incorporate complete corresponding information. 
Although the inconsistent background noise is not completely eliminated in the asymmetric window pair, due to the relatively small window space and the powerful relationship modeling ability of the correlation operation, the AWP is able to establish sufficient multi-modal correlations with less background noise interference.

To be specific, for each modality feature map $f_m^i \in {\mathbb{R}^{C \times H \times W}}$, we first partition it with a small pixel-level window of size M × M. Due to the spatial inconsistency, we divide the feature map of the other modality with a larger N × N window. Each larger window is obtained by expanding around the small window, aiming to cover the complete corresponding information. 
In this way, the feature maps with the small window are partitioned into $\left\lceil {\frac{H}{M}} \right\rceil  \times \left\lceil {\frac{W}{M}} \right\rceil $ non-overlapping windows. Since the position of the large window is determined by the corresponding small window, the number of large windows in the feature map is also $\left\lceil {\frac{H}{M}} \right\rceil  \times \left\lceil {\frac{W}{M}} \right\rceil $ but overlapped. 
Subsequently, the features within each large window are used to enhance those within the corresponding small one. 
For example, as shown in Fig.~\ref{fig::framework}, given a RGB feature map size of 8 × 8 with a window size of 2 × 2, the number of windows would be 4 × 4. To preserve the spacial consistency, the window size for the corresponding thermal feature map is 4 × 4, which is centered on the small RGB window.
Note that we perform the above operations on both modality features to ensure that each modality is enhanced by the asymmetric correlation, thus achieving a bi-directional cross-modal correlation. 

\subsubsection{Semantic Guidance Module (SGM)}
To focus cross-modal correlations on salient regions, we propose the SGM, which embeds semantic guidance into the correlation modeling. Since high-level features inherently contain rich semantic information that is capable of object category recognition and localization~\cite{mohammadi2020cagnet}, we utilize the high-level features of both modalities to enhance multi-level feature representations. However, the original high-level features of the two modalities also lack strong correlations and their modality complementarity is not explored, hindering accurate salient region localization. Therefore, the multi-modal high-level features need to be fully aggregated first. As shown at the bottom left of Fig.~\ref{fig::framework}, we collect the top extracted features ($f_{rgb}^4$ and $f_{t}^4$) as high-level features. Considering that the top features have large receptive fields and contain less noise, we directly use the correlation operation on their concatenated features. In this way, the intra-modal and inter-modal correlations can be established simultaneously to obtain a global multi-modal feature representation $f_{cat}^G$, which is formulated as:
\begin{equation}\label{Eq::cat}
	\begin{split}
		&f_{cat}^4 = [{\rm{flatten}}(f_{rgb}^4),{\rm{flatten}}(f_t^4)],
	\end{split}
\end{equation} 
\begin{equation}\label{Eq::corr1}
	\begin{split}
		&{f_{cat}^G} = \mathcal{C}(f_{cat}^4,f_{cat}^4),
	\end{split}
\end{equation} 
where $[,]$ represents the concatenation operation, and $f_{cat}^4$ is the high-level concatenated feature. $\mathcal{C}$ is the correlation operation:
\begin{equation}\label{Eq::att}
	\begin{split}
		&C(Q,V) = {\rm{softmax(}}\frac{{Q{K^T}}}{{\sqrt {{d_k}} }}{\rm{)}}V{\rm{ + }}Q,
	\end{split}
\end{equation} 
where $Q \in {\mathbb{R}^{{N_Q} \times {d_k}}}$, $K$, $V \in {\mathbb{R}^{{N_V} \times {d_k}}}$, and '$+$' represents the residual connection. The details can be seen at the bottom right of Fig.~\ref{fig::framework}, and we refer the reader to the literature~\cite{vaswani2017attention} for more detailed descriptions. Next, we split $f_{cat}^G$ along the spatial dimension into the global features of the two modalities, and reshape them into their initial shapes, denoted as $f_{rgb}^G$ and $f_{t}^G$. To match the spatial and channel dimensions of the features in different layers, we upsample them followed by a 3 × 3 convolutional layer. By multiplying the global feature with the feature maps to be partitioned, the salient regions in each window can be enhanced. This process can be formulated as:
\begin{equation}\label{Eq::enhance1}
	\begin{split}
		\hat f_{rgb}^i = Conv(U{p_{{2^{4 - i}}}}(f_{rgb}^G)) \odot f_{rgb}^i \hfill \\
	\end{split}
\end{equation} 
\begin{equation}\label{Eq::enhance2}
	\begin{split}
		\hat f_t^i = Conv(U{p_{{2^{4 - i}}}}(f_t^G)) \odot f_t^i, \hfill \\ 
	\end{split}
\end{equation} 
where $\hat{f}_{rgb}^i$ and $\hat{f}_{t}^i$ denote the enhanced features, $\odot$ is element-wise multiplication, $Conv$ is a convolution layer, $U{p_x}$ is the $x$× upsample operation with bilinear interpolation, and $i=2,...4$ means that we will not perform the correlation operation on the first-level features due to the heavy computational cost of high-resolution data. 

It is assumed that sequences from the small window of original RGB and thermal features are denoted as ${s_{rgb}} = [s_{rgb}^1,s_{rgb}^2,...,s_{rgb}^{M \times M}]$ and ${s_{t}} = [s_{t}^1,s_{t}^2,...,s_{t}^{M \times M}]$, and sequences from the corresponding large window of enhanced RGB and thermal features are denoted as ${\hat{s}_{rgb}} = [\hat{s}_{rgb}^1,\hat{s}_{rgb}^2,...,\hat{s}_{rgb}^{N \times N}]$ and ${\hat{s}_{t}} = [\hat{s}_{t}^1,\hat{s}_{t}^2,...,\hat{s}_{t}^{N \times N}]$. Therefore, the modality correlation modeled by ACM can be formulated as:
\begin{equation}\label{Eq::corr1}
	\begin{split}
		&\begin{gathered}
			{y_{rgb}} = C({s_{rgb}},\hat{{s_t}}) \hfill \\
			{y_t} = C({s_t},{\hat{s}_{rgb}}), \hfill \\ 
		\end{gathered}
	\end{split}
\end{equation} 
where $y_{rgb}$ and $y_{t}$ denote the output correlated sequences of the two modalities. $M$ and $N$ are the small window size and large window size, respectively. Through correlation operations on all asymmetric window pairs, strong cross-modal correlations for the whole salient region are established, and the feature representations of two modalities are thus improved, denoted as $\tilde{f}_{rgb}^i$ and $\tilde{f}_{t}^i$.


\begin{figure}[t]
	\centering
	\includegraphics[width=1\columnwidth]{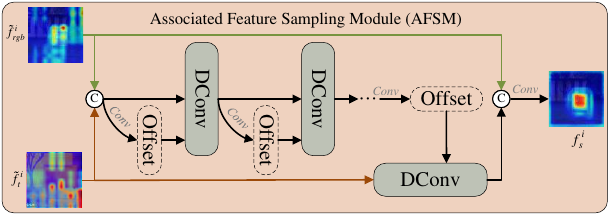}
	\caption{Details of the proposed associated feature sampling module} 
	\label{fig::AFSM}
\end{figure}
\subsection{Associated Feature Sampling Module (AFSM)}
\label{AFSM}
The features of the two modalities are correlated and enhanced by the ACM, but remain unaligned. Directly aggregating the unaligned features may lead to mismatches and inaccurate recognition for salient regions. To address this issue, we propose an Associated Feature Sampling Module (AFSM), which employs cascaded deformable convolutions (DConvs)~\cite{dai2017deformable} to sample relevant multi-modal features for further integration. The architecture of AFSM is shown in Fig.~\ref{fig::AFSM}. Here, we sample thermal features conditioned on corresponding RGB features. The AFSM first takes both RGB feature $\tilde{f}_{rgb}^i$ and thermal feature $\tilde{f}_{t}^i$ as inputs to predict sampling offsets $\alpha^i$ for the thermal feature $\tilde{f}_{t}^i$:
\begin{equation}\label{Eq::para}
	\begin{split}
		&\alpha^i  = {f_\alpha^i }(\tilde{f}_t^i,\tilde{f}_{rgb}^i) = \left\{ {\Delta {p_n}|n = 1,...,|R|} \right\},
	\end{split}
\end{equation} 
where $R = \{ ( - 1, - 1),( - 1,0),...,(0,1),(1,1)\} $ represents a standard grid of a 3 × 3 kernel, $f_\alpha^i$ denotes the offset generation function. Next, we take $\alpha^i$ and $\tilde{f}_{t}^i$ as inputs for deformable convolution to compute a newly sampled thermal feature $\tilde{f}_{t \to rgb}^i$. Each position ${p_0}$ on the sampled thermal feature map can be formulated as:
\begin{equation}\label{Eq::sample}
	\begin{split}
		&\tilde{f}_{t \to rgb}^i({p_0}) = \sum\limits_{{p_n} \in R} {\omega ({p_n}){\tilde{f}_t^i}({p_0} + {p_n} + \Delta {p_n})} ,
	\end{split}
\end{equation} 
where $\Delta {p_n}$ is a learnable offset, which may be fractional and thus computed through bilinear interpolation, more details can be found in~\cite{dai2017deformable}. In practice, our AFSM cascades four deformable convolutional layers to sample more relevant and accurate saliency features in the thermal modality. Then, the sampled thermal features and corresponding RGB features are concatenated along the channel dimension, and fed into a convolutional layer with a 3 × 3 kernel to aggregate their saliency information, as follows:
\begin{equation}\label{Eq::integration}
	\begin{split}
		&f_s^i = Conv([\tilde{f}_{t \to rgb}^i,\tilde{f}_{rgb}^i]) ,
	\end{split}
\end{equation} 
where $f_s^i$ ($i=2,...,4$) denotes the integrated multi-modal features, which will be aggregated for final prediction.
\begin{table*}[t]
	\centering
	\caption{Comparison of UVT2000 With the Prevalent Multi-modal SOD Datasets on Modality, Scene Number, Category Number, Collection Method, Manual Alignment, Challenge Annotation, and Resolution.}
	\resizebox{1\textwidth}{!}{
		\begin{tabular}{c|c|c|ccc|c|c|c|c|c|ccc}
			\toprule
			\multicolumn{2}{c|}{\multirow{2}[4]{*}{Dataset}} & \multirow{2}[4]{*}{Year} & \multicolumn{3}{c|}{Modality} & Scene & Category & Collection & Manual & Challenge & \multicolumn{3}{c}{Resolution} \\
			\cmidrule{4-6}\cmidrule{12-14}    \multicolumn{2}{c|}{} &       & RGB   & Thermal & Depth & Number & Number & Method & Alignment & Annotation & RGB   & Thermal & Depth \\
			\midrule
			\multirow{6}[2]{*}{Aligned} & VT821~\cite{wang2018rgb} & 2018  & 821   & 821   & -     & 16 & 198 & Camera & \checkmark     & \checkmark     & 640 $\times$ 480 & 640 $\times$ 480 & - \\
			& VT1000~\cite{tu2019rgb} & 2019  & 1000  & 1000  & -     & 76 & 329 & Camera & \checkmark     & \checkmark     & 640 $\times$ 480 & 640 $\times$ 480 & - \\
			& VT5000~\cite{tu2020rgbt} & 2020  & 5000  & 5000  & -     & 212 & 304 & Camera & \checkmark     & \checkmark     & 640 $\times$ 480 & 640 $\times$ 480 & - \\
			& NJUD~\cite{ju2014depth}  & 2014  & 1985  & -     & 1985  & 282 & 359 & Movie/Internet & \checkmark     & $\times$     & 256 $\times$ 256 & -     & 256 $\times$ 256 \\
			& DUT-RGBD~\cite{piao2019depth} & 2019  & 1200  & -     & 1200  & 191 & 291 & Camera & \checkmark     & $\times$     & 256 $\times$ 256 & -     & 256 $\times$ 256 \\
			& SIP~\cite{fan2020rethinking}   & 2020  & 929   & -     & 929   & 69 & 1 & Mobile phone & \checkmark     & $\times$     & 256 $\times$ 256 & -     & 256 $\times$ 256 \\
			\midrule
			Weakly & un-VT821~\cite{tu2022weakly} & 2022  & 821   & 821   & -     & 16 & 198 & Augmentation & \checkmark     & \checkmark     & 640 $\times$ 480 & 640 $\times$ 480 & - \\
			Aligned & un-VT1000~\cite{tu2022weakly} & 2022  & 1000  & 1000  & -     & 76 & 329 &  of & \checkmark     & \checkmark     & 640 $\times$ 480 & 640 $\times$ 480 & - \\
			& un-VT5000~\cite{tu2022weakly} & 2022  & 5000  & 5000  & -     & 212 & 304 & existing datasets & \checkmark     & \checkmark     & 640 $\times$ 480 & 640 $\times$ 480 & - \\
			\midrule
			Unaligned & UVT2000 & -     & 2000  & 2000  & -     & 295 & 429 & Camera & $\times$     & \checkmark     & 2048 $\times$ 1536 & 640 $\times$ 480 & - \\
			\bottomrule
		\end{tabular}%
	}
	\label{tab:dataset_comp}%
\end{table*}%

\begin{figure*}[t]
	\centering
	\includegraphics[width=1\textwidth]{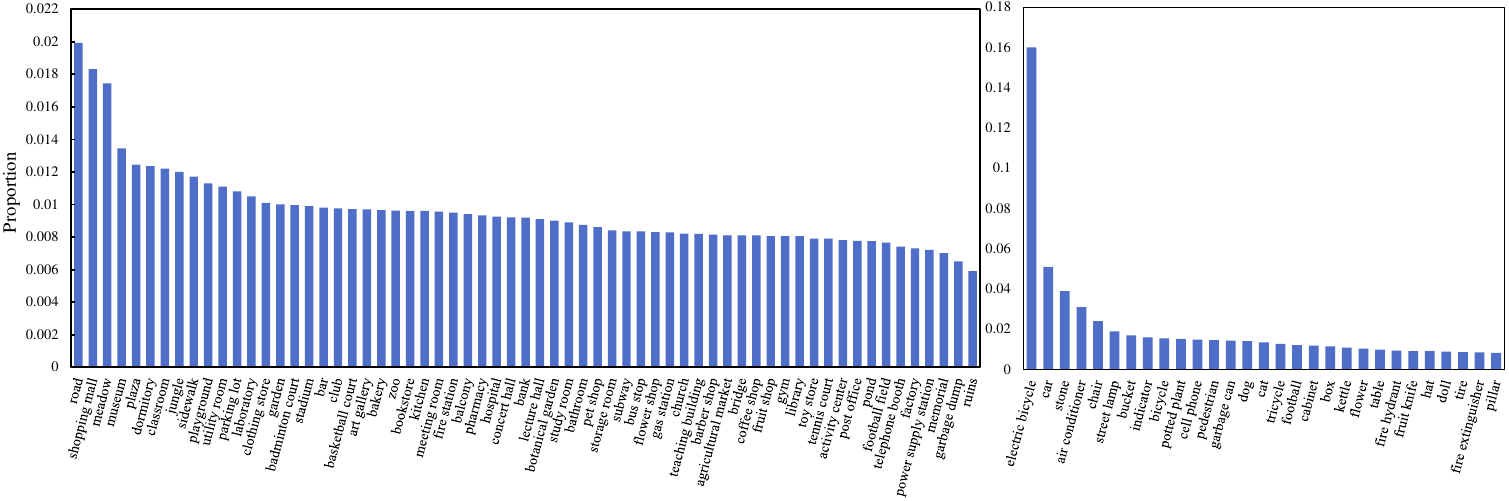}
	\caption{Top 60\% scene and object category distributions in our proposed UVT2000 dataset.} 
	\label{fig::scene}
\end{figure*}
\subsection{Saliency Prediction and Loss Function}
\label{loss}
Similar to U-Net framework~\cite{ronneberger2015u}, the decoding process for saliency map prediction is formed through top-down feature fusion, which can be formulated as:
\begin{equation}\label{Eq::topdown}
	\begin{split}
		&\hat{f}_s^i = \left\{ \begin{gathered}
			Conv(U{p_2}(f_s^i + \hat{f}_s^{i + 1}),i = 2,3 \hfill \\
			Conv(U{p_2}(f_s^i),i = 4 \hfill \\ 
		\end{gathered}  \right.
	\end{split}
\end{equation} 
\begin{equation}\label{Eq::predition}
	\begin{split}
		&S = Conv(U{p_4}(Conv(\hat{f}_s^2 + f_{rgb}^1 + f_t^1))),
	\end{split}
\end{equation} 
where $S$ denotes the final predicted saliency map. Note that due to the large computational cost of high-resolution feature maps, the first-level features ($f_{rgb}^1$ and $f_t^1$) are only introduced here to supplement detailed information.

Following ~\cite{tu2021multi,wu2021mobilesal}, we use a combination of binary cross-entropy loss, smoothness loss~\cite{godard2017unsupervised}, and dice loss~\cite{milletari2016v} to optimize the proposed method, which can be described as:

\begin{equation}\label{Eq::loss}
	\begin{split}
		&{\cal L} = {{\cal L} _{bce}}(S,G) + {{\cal L} _{smooth}}(S,G) + {{\cal L} _{dice}}(S,G),
	\end{split}
\end{equation} 
where ${\cal L}$ represents the loss function, and $G$ denotes the ground truth.

\section{UVT2000 Benchmark}

Existing RGBT SOD datasets~\cite{wang2018rgb,tu2019rgb,tu2020rgbt} are manually aligned, which is labor-intensive and limits the research on alignment-free RGBT SOD. Although DCNet~\cite{tu2022weakly} builds weakly aligned datasets (i.e., unaligned-VT5000~\cite{tu2022weakly}, unaligned-VT1000~\cite{tu2022weakly}, unaligned-VT821~\cite{tu2022weakly}) by performing random affine transformation for the image pairs in existing datasets (i.e., VT5000, VT1000, and VT821), the misalignment of these image pairs is artificial and weak with a lack of practical significance. 
To facilitate the research on alignment-free RGBT SOD, we construct a benchmark dataset with 2000 unaligned visible-thermal image pairs, denoted as UVT2000.

\subsection{Data Acquisition}
UVT2000 dataset is captured by a FLIR SC620 equipped with a pair of CCD and thermal infrared cameras, the same as used in~\cite{tu2019rgb}. Due to parallax and different size viewing angles between the two cameras, the same object in their captured image pairs inherently suffers from positional shifts and scale differences, as shown in Fig.~\ref{fig::camera}. Therefore, the degree of misalignment in UVT2000 derives from actual camera shots, and reflects the practical situations in real-world applications. UVT2000 also avoids a number of labor-intensive operations such as manual cropping and rescaling.

Given that visible images are more in line with human visual preferences, we capture these image pairs mainly based on RGB modality, except for visible image degradation cases, such as low illumination and out of focus. To ensure the quality and reliability of the dataset, we initially collect about 2500 pairs of images for selection.

\subsection{Dataset Annotation}
Before annotation, we first discard obviously low-quality image pairs, such as image pairs with no objects. Then, five annotators are asked to select the objects in each image pair that are salient at first glance. Based on the agreement of the selected salient objects, we rank these image pairs. Eventually, we annotate the highest-ranking 2000 image pairs. Since the data collection process is mainly based on the RGB modality, we annotate pixel-level ground truth according to visible images supplemented by corresponding thermal images.

\begin{figure*}[t]
	\centering
	\includegraphics[width=1\textwidth]{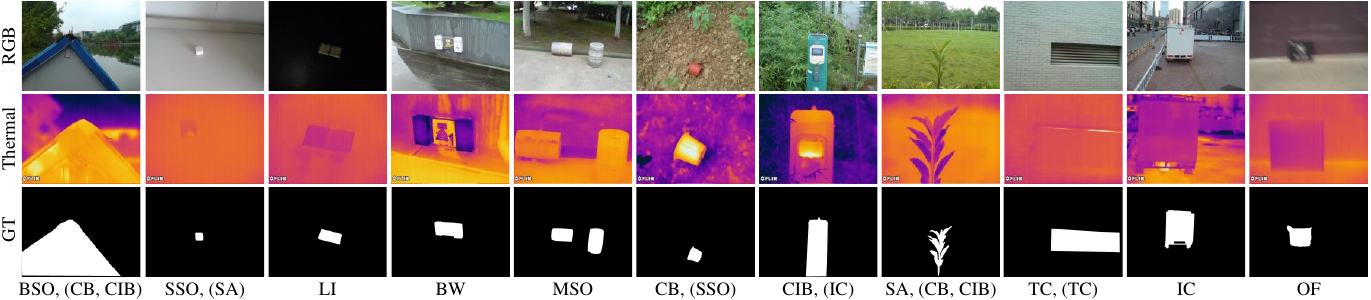}
	\caption{Examples of visible-thermal image pairs with ground truth and challenge annotations in the proposed benchmark dataset UVT2000. GT: ground truth; BSO: big salient object; SSO: small salient object; LI: low illumination; BW: bad weather; MSO: multiple salient objects; CB: center bias; CIB: cross image boundary; SA: similar appearance; TC: thermal crossover; IC: image clutter; OF: out of focus. A, (B) indicates that the image pair contains the main challenge A and other secondary challenge(s) B.} 
	\label{fig::camera}
\end{figure*}

\begin{figure}[t]
	\centering
	\includegraphics[width=1\columnwidth]{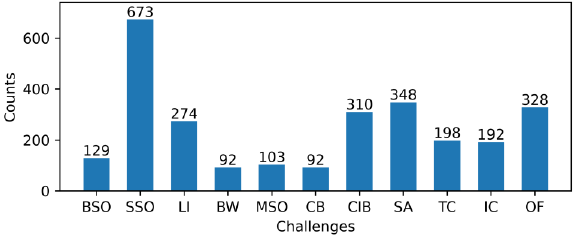}
	\caption{Challenge distribution of the newly constructed dataset UVT2000.} 
	\label{fig::tongji}
\end{figure}

The image pairs in our dataset are captured in various scenes. We annotate them with 11 challenges to facilitate different methods to address these challenges. These challenges are: big salient object (BSO), small salient object (SSO), low illumination (LI), bad weather (BW), multiple salient objects (MSO), center bias (CB), cross image boundary (CIB), similar appearance (SA), thermal crossover (TC), image clutter (IC), and out of focus (OF). More detailed descriptions of these challenges can be found in~\cite{tu2019rgb}. Fig.~\ref{fig::camera} shows the examples under different challenges. Note that each image pair may be annotated with multiple challenge attributes, we only describe the main challenge for each example here. Fig.~\ref{fig::tongji} also illustrates the challenge distribution of the UVT2000. 

In particular, the image pairs of each scene in Fig.~\ref{fig::camera} appear to suffer from center bias, but center bias is not a common challenge in the UVT2000 dataset for the following reasons. First, although objects in the thermal modality are generally shifted from the image center due to unalignment, the challenge annotations are mainly based on the RGB modality, which is not always shifted from the image center. Second, as defined in~\cite{tu2019rgb}, center bias refers to the center of salient objects being far away from the image center. Although the salient objects in some examples appear to be close to the image boundary, their centers are still close to the image center, so they are not grouped into the center bias challenge, such as the examples in the seventh and ninth columns in Fig.~\ref{fig::camera}.

\subsection{Dataset Characteristics}
As shown in Table~\ref{tab:dataset_comp}, compared with existing prevalent multi-modal SOD datasets, UVT2000 mainly has the following characteristics:

1) \textbf{No manual alignment is required for constructing UVT2000.} 
Since UVT2000 is proposed to facilitate the study of alignment-free RGBT SOD, the image pairs are captured directly by the camera without any manual alignment operations.

2) \textbf{The UVT2000 is more diverse in scenes and categories.} 
Based on~\cite{liu2021learning}, Table~\ref{tab:dataset_comp} reports the total number of scenes and categories for UVT2000 and the compared datasets. The results show that UVT2000 has more diverse scenes and categories. Fig.~\ref{fig::scene} further illustrates the distribution of the top 60\% scenes and object categories in the proposed UVT2000. It can be seen that most scenes and categories have approximately smooth distributions.

3) \textbf{UVT2000 is more challenging and practical.} 
As the first unaligned RGBT SOD dataset, compared to existing RGBT SOD datasets, UVT2000 is of smaller scale than VT5000 and unaligned-VT5000 (i.e., a variant of VT5000), while it is more consistent with actual scenes and presents greater challenges. As shown in Table~\ref{tab:dataset_comp}, compared with the weakly aligned datasets, UVT2000 is collected through real-world camera shooting, which has more practical significance.

4) \textbf{High resolution of visible images is maintained in UVT2000.} 
With no rescaling and cropping required for alignment, the UVT2000 maintains the original resolution of image pairs with high quality.
%

\section{Experiments}
\begin{table*}[t]
	\centering
	\caption{Quantitative Comparison of E-measure (${E_\xi }$), S-measure (${S_\alpha }$), Weighed F-measure (${F_\beta ^\omega}$), and Mean Absolute Error ($MAE$) on One Unaligned, Three Weakly Aligned, and Three Aligned Datasets. The Best Three Results Are Marked with \textcolor[rgb]{ .792,  0,  0}{\textbf{Red}}, \textcolor[rgb]{ 0,  .533,  .2}{\textbf{Green}}, and \textcolor[rgb]{ 0,  .4,  .8}{\textbf{Blue}}.}
	\resizebox{1\textwidth}{!}{
		\begin{tabular}{cc|cccccccccccccc|cc}
			\toprule
			\multicolumn{2}{c|}{\multirow{2}[2]{*}{Method}} & ADF$_{20}$ & MIDD$_{21}$ & CSRNet$_{21}$ & CGFNet$_{21}$ & SwinNet$_{22}$ & OSRNet$_{22}$ & TNet$_{22}$ & DCNet$_{22}$ & MCFNet$_{23}$ & HRTransNet$_{23}$ & LSNet$_{23}$ & CAVER$_{23}$ & WaveNet$_{23}$ & SPNet$_{23}$ & SACNet & SACNet \\
			\multicolumn{2}{c|}{} &\cite{tu2020rgbt}      &\cite{tu2021multi}       &\cite{huo2021efficient}       &\cite{wang2021cgfnet}       &\cite{liu2021swinnet}       &\cite{huo2022real}       &\cite{TNet}       &\cite{tu2022weakly}       &\cite{ma2023modal}       &\cite{Tang2023RTransNet}       &\cite{zhou2023lsnet}       &\cite{pang2023caver}       &\cite{zhou2023wave}       &\cite{zhang23SPNet}       & Ours  & Ours \\
			\midrule
			\multicolumn{2}{c|}{Backbone} & VGG16 & VGG16 & ESPNetv2 & VGG16 & SwinB & VGG16 & ResNet50 & VGG16 & ResNet50 & HRFormer & MobileNet-v2 & ResNet50 & Wave-MLP & PVT-v2 & ResNet50 & SwinB \\
			\midrule
			\multicolumn{2}{c|}{FPS \hfill$\uparrow$} & 27    & 33    & 75    & 18    & 34    & \textcolor[rgb]{ 0,  .533,  .2}{\textbf{142}} & \textcolor[rgb]{ 0,  .4,  .8}{\textbf{81}} & 43    & 72    & 37    & \textcolor[rgb]{ .792,  0,  0}{\textbf{314}} & 67    & 17    & 29    & 19    & 27 \\
			\multicolumn{2}{c|}{Parameters(M) \hfill$\downarrow$} & 66.8  & 52.4  & \textcolor[rgb]{ .792,  0,  0}{\textbf{1.0}} & 66.4  & 199.2  & \textcolor[rgb]{ 0,  .4,  .8}{\textbf{15.6}} & 87.0  & 24.1  & 70.8  & 26.3  & \textcolor[rgb]{ 0,  .533,  .2}{\textbf{4.6}} & 55.8  & 80.7  & 110.0  & 530.9  & 327.7  \\
			\midrule
			\multirow{5}[1]{*}{\begin{sideways}UVT2000\end{sideways}} & ${E_\xi}$ \hfill\hfill$\uparrow$  & 0.640  & 0.673  & 0.605  & 0.705  & \textcolor[rgb]{ 0,  .4,  .8}{\textbf{0.743}} & 0.732  & 0.717  & 0.753  & 0.727  & 0.603  & 0.679  & 0.727  & 0.634  & 0.733 & \textcolor[rgb]{ 0,  .533,  .2}{\textbf{0.777}} & \textcolor[rgb]{ .792,  0,  0}{\textbf{0.812}} \\
			& ${S_\alpha}$ \hfill\hfill$\uparrow$  & 0.672  & 0.721  & 0.629  & 0.740  & \textcolor[rgb]{ 0,  .533,  .2}{\textbf{0.777}} & 0.696  & 0.754  & 0.731  & 0.739  & 0.651  & 0.728  & 0.749  & 0.699  & \textcolor[rgb]{ 0,  .4,  .8}{\textbf{0.765}} & 0.759 & \textcolor[rgb]{ .792,  0,  0}{\textbf{0.807}} \\
			& ${F_\beta ^\omega}$ \hfill\hfill$\uparrow$  & 0.349  & 0.445  & 0.349  & 0.477  & \textcolor[rgb]{ 0,  .4,  .8}{\textbf{0.551}} & 0.454  & 0.527  & 0.513  & 0.464  & 0.375  & 0.466  & 0.527  & 0.399  & \textcolor[rgb]{ 0,  .533,  .2}{\textbf{0.558}} & 0.544 & \textcolor[rgb]{ .792,  0,  0}{\textbf{0.640}} \\
			& $MAE$\hfill$\downarrow$ & 0.078  & 0.075  & 0.116  & 0.067  & \textcolor[rgb]{ 0,  .4,  .8}{\textbf{0.051}} & 0.051  & 0.058  & 0.049  & 0.066  & 0.116  & 0.065  & 0.061  & 0.078  & 0.059 & \textcolor[rgb]{ 0,  .533,  .2}{\textbf{0.038}} & \textcolor[rgb]{ .792,  0,  0}{\textbf{0.036}} \\
			\midrule
			\multirow{5}[1]{*}{\begin{sideways}VT5000\end{sideways}} & ${E_\xi}$ \hfill\hfill$\uparrow$  & 0.891  & 0.897  & 0.905  & 0.922  & 0.942  & 0.908  & 0.927  & 0.920  & 0.924  & \textcolor[rgb]{ 0,  .4,  .8}{\textbf{0.945}} & 0.915  & 0.924  & 0.940  & \textcolor[rgb]{ 0,  .533,  .2}{\textbf{0.948}} & 0.933  & \textcolor[rgb]{ .792,  0,  0}{\textbf{0.957}} \\
			& ${S_\alpha}$ \hfill\hfill$\uparrow$  & 0.864  & 0.868  & 0.868  & 0.883  & \textcolor[rgb]{ 0,  .4,  .8}{\textbf{0.912}} & 0.875  & 0.895  & 0.871  & 0.887  & \textcolor[rgb]{ 0,  .4,  .8}{\textbf{0.912}} & 0.877  & 0.892  & 0.911  & \textcolor[rgb]{ 0,  .533,  .2}{\textbf{0.914}} & 0.892  & \textcolor[rgb]{ .792,  0,  0}{\textbf{0.917}} \\
			& ${F_\beta ^\omega}$ \hfill\hfill$\uparrow$  & 0.722  & 0.763  & 0.796  & 0.831  & 0.846  & 0.807  & 0.840  & 0.819  & 0.836  & \textcolor[rgb]{ 0,  .4,  .8}{\textbf{0.870}} & 0.806  & 0.835  & 0.864  & \textcolor[rgb]{ 0,  .533,  .2}{\textbf{0.880}} & 0.838  & \textcolor[rgb]{ .792,  0,  0}{\textbf{0.888}} \\
			& $MAE$\hfill$\downarrow$ & 0.048  & 0.043  & 0.042  & 0.035  & 0.026  & 0.040  & 0.033  & 0.035  & 0.033  & \textcolor[rgb]{ 0,  .4,  .8}{\textbf{0.025}} & 0.037  & 0.032  & 0.026  & \textcolor[rgb]{ 0,  .533,  .2}{\textbf{0.024}} & 0.030  & \textcolor[rgb]{ .792,  0,  0}{\textbf{0.021}} \\
			\midrule
			\multirow{5}[1]{*}{\begin{sideways}~~un-VT5000\end{sideways}} & ${E_\xi}$ \hfill\hfill$\uparrow$  & 0.824  & 0.885  & 0.804  & 0.899  & 0.923  & 0.770  & 0.910  & 0.908  & 0.905  & 0.847  & 0.890  & 0.917  & 0.831  & \textcolor[rgb]{ 0,  .533,  .2}{\textbf{0.929}} & \textcolor[rgb]{ 0,  .4,  .8}{\textbf{0.925}} & \textcolor[rgb]{ .792,  0,  0}{\textbf{0.949}} \\
			& ${S_\alpha}$ \hfill\hfill$\uparrow$  & 0.813  & 0.854  & 0.746  & 0.865  & \textcolor[rgb]{ 0,  .4,  .8}{\textbf{0.899}} & 0.724  & 0.879  & 0.854  & 0.864  & 0.811  & 0.856  & 0.884  & 0.825  & \textcolor[rgb]{ 0,  .533,  .2}{\textbf{0.900}} & 0.879  & \textcolor[rgb]{ .792,  0,  0}{\textbf{0.911}} \\
			& ${F_\beta ^\omega}$ \hfill\hfill$\uparrow$  & 0.625  & 0.740  & 0.602  & 0.746  & \textcolor[rgb]{ 0,  .4,  .8}{\textbf{0.823}} & 0.571  & 0.806  & 0.790  & 0.757  & 0.692  & 0.757  & 0.822  & 0.664  & \textcolor[rgb]{ 0,  .533,  .2}{\textbf{0.848}} & 0.820  & \textcolor[rgb]{ .792,  0,  0}{\textbf{0.876}} \\
			& $MAE$\hfill$\downarrow$  & 0.072  & 0.049  & 0.089  & 0.046  & \textcolor[rgb]{ 0,  .4,  .8}{\textbf{0.031}} & 0.106  & 0.038  & 0.041  & 0.044  & 0.068  & 0.046  & 0.038  & 0.057  & \textcolor[rgb]{ 0,  .533,  .2}{\textbf{0.030}} & 0.034  & \textcolor[rgb]{ .792,  0,  0}{\textbf{0.023}} \\
			\midrule
			\multirow{5}[1]{*}{\begin{sideways}VT1000\end{sideways}} & ${E_\xi}$ \hfill\hfill$\uparrow$  & 0.921  & 0.933  & 0.925  & 0.944  & 0.947  & 0.935  & 0.937  & 0.948  & 0.944  & 0.945  & 0.935  & 0.945  & \textcolor[rgb]{ 0,  .4,  .8}{\textbf{0.952}} & \textcolor[rgb]{ 0,  .533,  .2}{\textbf{0.954}} & 0.949  & \textcolor[rgb]{ .792,  0,  0}{\textbf{0.958}} \\
			& ${S_\alpha}$ \hfill\hfill$\uparrow$  & 0.910  & 0.915  & 0.918  & 0.923  & 0.938  & 0.926  & 0.929  & 0.922  & 0.932  & 0.938  & 0.925  & 0.936  & \textcolor[rgb]{ .792,  0,  0}{\textbf{0.945}} & \textcolor[rgb]{ 0,  .4,  .8}{\textbf{0.941}} & 0.932  & \textcolor[rgb]{ 0,  .533,  .2}{\textbf{0.942}} \\
			& ${F_\beta ^\omega}$ \hfill\hfill$\uparrow$  & 0.804  & 0.856  & 0.878  & 0.900  & 0.894  & 0.891  & 0.895  & 0.902  & 0.906  & 0.913  & 0.887  & 0.909  & \textcolor[rgb]{ 0,  .4,  .8}{\textbf{0.921}} & \textcolor[rgb]{ 0,  .533,  .2}{\textbf{0.925}} & 0.907  & \textcolor[rgb]{ .792,  0,  0}{\textbf{0.927}} \\
			& $MAE$\hfill$\downarrow$ & 0.034  & 0.027  & 0.024  & 0.023  & 0.018  & 0.022  & 0.021  & 0.021  & 0.019  & \textcolor[rgb]{ 0,  .4,  .8}{\textbf{0.017}} & 0.023  & \textcolor[rgb]{ 0,  .4,  .8}{\textbf{0.017}} & \textcolor[rgb]{ 0,  .533,  .2}{\textbf{0.015}} & \textcolor[rgb]{ 0,  .533,  .2}{\textbf{0.015}} & 0.018  & \textcolor[rgb]{ .792,  0,  0}{\textbf{0.014}} \\
			\midrule
			\multirow{5}[1]{*}{\begin{sideways}~~un-VT1000\end{sideways}} & ${E_\xi}$ \hfill\hfill$\uparrow$  & 0.876  & 0.919  & 0.853  & 0.922  & 0.938  & 0.825  & 0.927  & 0.943  & 0.929  & 0.891  & 0.919  & \textcolor[rgb]{ 0,  .4,  .8}{\textbf{0.940}} & 0.863  & 0.938  & \textcolor[rgb]{ 0,  .533,  .2}{\textbf{0.944}} & \textcolor[rgb]{ .792,  0,  0}{\textbf{0.954}} \\
			& ${S_\alpha}$ \hfill\hfill$\uparrow$  & 0.873  & 0.904  & 0.817  & 0.914  & \textcolor[rgb]{ 0,  .533,  .2}{\textbf{0.936}} & 0.800  & 0.920  & 0.915  & 0.914  & 0.879  & 0.910  & \textcolor[rgb]{ 0,  .4,  .8}{\textbf{0.932}} & 0.875  & 0.931  & 0.925 & \textcolor[rgb]{ .792,  0,  0}{\textbf{0.941}} \\
			& ${F_\beta ^\omega}$ \hfill\hfill$\uparrow$  & 0.735  & 0.830  & 0.730  & 0.833  & 0.890  & 0.701  & 0.877  & 0.889  & 0.833  & 0.810  & 0.853  & \textcolor[rgb]{ 0,  .533,  .2}{\textbf{0.902}} & 0.758  & \textcolor[rgb]{ 0,  .533,  .2}{\textbf{0.902}} & \textcolor[rgb]{ 0,  .4,  .8}{\textbf{0.897}} & \textcolor[rgb]{ .792,  0,  0}{\textbf{0.923}} \\
			& $MAE$\hfill$\downarrow$ & 0.051  & 0.033  & 0.069  & 0.031  & \textcolor[rgb]{ 0,  .533,  .2}{\textbf{0.018}} & 0.077  & 0.025  & 0.023  & 0.028  & 0.045  & 0.028  & 0.020  & 0.042  & \textcolor[rgb]{ 0,  .4,  .8}{\textbf{0.019}} & 0.021 & \textcolor[rgb]{ .792,  0,  0}{\textbf{0.014}} \\
			\midrule
			\multirow{5}[2]{*}{\begin{sideways}VT821\end{sideways}} & ${E_\xi}$ \hfill\hfill$\uparrow$  & 0.842  & 0.895  & 0.909  & 0.912  & 0.926  & 0.896  & 0.919  & 0.912  & 0.918  & \textcolor[rgb]{ 0,  .4,  .8}{\textbf{0.929}} & 0.911  & 0.919  & \textcolor[rgb]{ 0,  .4,  .8}{\textbf{0.929}} & \textcolor[rgb]{ .792,  0,  0}{\textbf{0.936}} & 0.917  & \textcolor[rgb]{ 0,  .533,  .2}{\textbf{0.932}} \\
			& ${S_\alpha}$ \hfill\hfill$\uparrow$  & 0.810  & 0.871  & 0.884  & 0.881  & 0.904  & 0.875  & 0.899  & 0.876  & 0.891  & \textcolor[rgb]{ 0,  .4,  .8}{\textbf{0.906}} & 0.878  & 0.891  & \textcolor[rgb]{ 0,  .533,  .2}{\textbf{0.912}} & \textcolor[rgb]{ .792,  0,  0}{\textbf{0.913}} & 0.883  & \textcolor[rgb]{ 0,  .4,  .8}{\textbf{0.906}} \\
			& ${F_\beta ^\omega}$ \hfill\hfill$\uparrow$  & 0.627  & 0.760  & 0.821  & 0.829  & 0.818  & 0.801  & 0.841  & 0.823  & 0.835  & 0.849  & 0.809  & 0.835  & \textcolor[rgb]{ 0,  .533,  .2}{\textbf{0.863}} & \textcolor[rgb]{ .792,  0,  0}{\textbf{0.873}} & 0.817  & \textcolor[rgb]{ 0,  .4,  .8}{\textbf{0.859}} \\
			& $MAE$\hfill$\downarrow$ & 0.716  & 0.045  & 0.038  & 0.038  & 0.030  & 0.043  & 0.030  & 0.033  & 0.029  & 0.026  & 0.033  & 0.033  & \textcolor[rgb]{ 0,  .533,  .2}{\textbf{0.024}} & \textcolor[rgb]{ .792,  0,  0}{\textbf{0.023}} & 0.033  & \textcolor[rgb]{ 0,  .4,  .8}{\textbf{0.025}} \\
			\midrule
			\multirow{5}[1]{*}{\begin{sideways}~un-VT821\end{sideways}} & ${E_\xi}$ \hfill\hfill$\uparrow$  & 0.818  & 0.888  & 0.801  & 0.875  & 0.905  & 0.790  & 0.889  & 0.908  & 0.899  & 0.873  & 0.888  & 0.887  & 0.843  & \textcolor[rgb]{ 0,  .4,  .8}{\textbf{0.910}} & \textcolor[rgb]{ 0,  .533,  .2}{\textbf{0.913}} & \textcolor[rgb]{ .792,  0,  0}{\textbf{0.929}} \\
			& ${S_\alpha}$ \hfill\hfill$\uparrow$  & 0.800  & 0.866  & 0.750  & 0.854  & \textcolor[rgb]{ 0,  .4,  .8}{\textbf{0.888}} & 0.733  & 0.873  & 0.860  & 0.867  & 0.839  & 0.852  & 0.870  & 0.826  & \textcolor[rgb]{ 0,  .533,  .2}{\textbf{0.894}} & 0.869 & \textcolor[rgb]{ .792,  0,  0}{\textbf{0.905}} \\
			& ${F_\beta ^\omega}$ \hfill\hfill$\uparrow$  & 0.616  & 0.747  & 0.605  & 0.736  & \textcolor[rgb]{ 0,  .4,  .8}{\textbf{0.799}} & 0.575  & 0.788  & 0.799  & 0.741  & 0.736  & 0.746  & 0.795  & 0.670  & \textcolor[rgb]{ 0,  .533,  .2}{\textbf{0.833}} & 0.789 & \textcolor[rgb]{ .792,  0,  0}{\textbf{0.857}} \\
			& $MAE$\hfill$\downarrow$ & 0.073  & 0.048  & 0.089  & 0.063  & \textcolor[rgb]{ 0,  .4,  .8}{\textbf{0.036}} & 0.086  & 0.047  & 0.036  & 0.044  & 0.054  & 0.044  & \textcolor[rgb]{ 0,  .4,  .8}{\textbf{0.036}} & 0.056  & \textcolor[rgb]{ 0,  .533,  .2}{\textbf{0.033}} & 0.037 & \textcolor[rgb]{ .792,  0,  0}{\textbf{0.026}} \\
			\bottomrule
		\end{tabular}%
	}
	\label{tab:compare}%
\end{table*}%

\subsection{Datasets}
To evaluate the effectiveness of our method, we conduct experiments  on three aligned datasets, three weakly aligned datasets, and one unaligned dataset, including VT821~\cite{wang2018rgb}, VT1000~\cite{tu2019rgb}, VT5000~\cite{tu2020rgbt}, unaligned-VT821~\cite{tu2022weakly} (i.e., un-VT821), unaligned-VT1000~\cite{tu2022weakly} (i.e., un-VT1000), unaligned-VT5000~\cite{tu2022weakly} (i.e., un-VT5000), and the proposed UVT2000. VT821 contains 821 aligned image pairs, some of which are added with noise to make the dataset more challenging. VT1000 includes 1000 pairs of visible-thermal images that are aligned and collected in relatively simple scenes. VT5000 consists of 5000 aligned image pairs with a variety of object sizes and scenes, which are divided in half into a training set and a testing set.
The un-VT821, un-VT1000, and un-VT5000 are weakly aligned datasets obtained by performing random affine transformations on the corresponding aligned datasets above. Following ~\cite{tu2022weakly}, we utilize the training sets of VT5000 to train the model for aligned datasets, and train the model with the training sets of un-VT5000 for weakly aligned and unaligned datasets. The remaining datasets are used for testing.

\subsection{Evaluation Metrics}
We adopt four widely used evaluation metrics for evaluation, including E-measure (${E_\xi}$), S-measure (${S_\alpha }$), weighed F-measure (${F_\beta ^\omega}$), and mean absolute error ($MAE$). Specifically, E-measure measures both image-level statistics and pixel-level matching information. S-measure evaluates the structural similarity at the region and object level. Weighed F-measure is a weighted combination of precision and recall. Mean absolute error indicates the absolute error between predictions and ground truth. We also introduce the "precision-recall" curve to demonstrate the overall performance of the model. To assess model complexity, we also report the results in terms of FPS (Frame-Per-Second) and number of parameters.

\subsection{Implementation Details}
Our framework is implemented with Pytorch in a workspace with two RTX 3090 GPUs. Input images are resized into 384 × 384 for both training and testing. During the training stage, we apply the AdamW algorithm with a learning rate of 1e-5 and a weight decay of 1e-4 to optimize our network. 
We set the batch size as 8 and epoch as 200 to train all our models, which takes about 15 hours. The backbone of our network is the SwinB network~\cite{liu2021swin} pre-trained on ImageNet. In our network, the small window size and large window size are set to 4 and 6, respectively.

\begin{figure*}[t]
	\centering
	\includegraphics[width=1\linewidth]{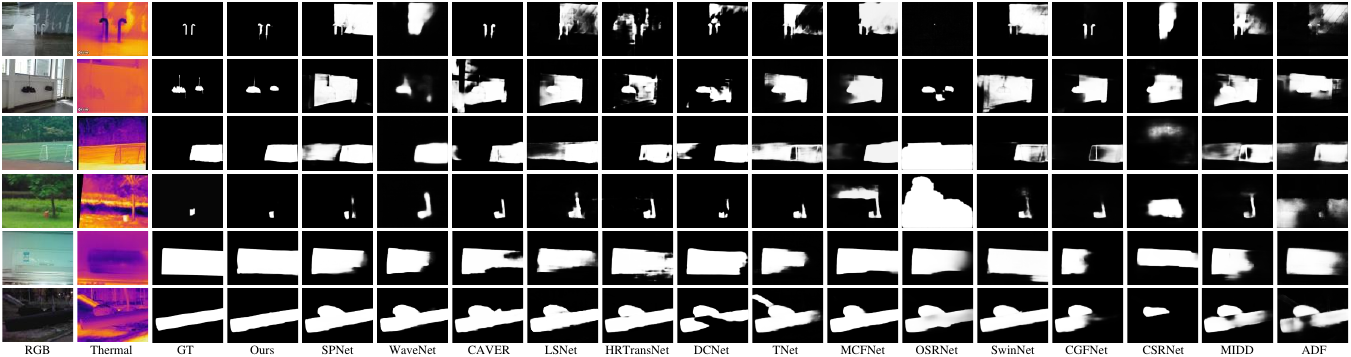}
	\caption{Visual comparisons with other state-of-the-art methods in some challenging scenes, including bad weather (e.g., Row 1), multiple salient objects (e.g., Row 2), center bias (e.g., Row 3), small salient object (e.g., Row 4), similar foreground and background (e.g., Row 5), and low illumination (e.g., Row 6).} 
	\label{fig::dingxing}
\end{figure*}
\begin{figure}[t]
	\centering
	\includegraphics[width=1\columnwidth]{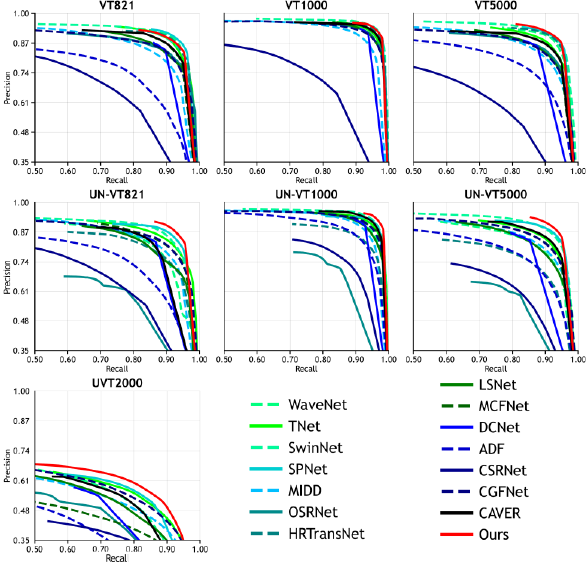}
	\caption{The precision-recall (PR) curves of our method and 14 compared methods.} 
	\label{fig::pr}
\end{figure}

\subsection{Comparison with State-of-the-arts}
We compare our method with 14 state-of-the-art RGBT SOD methods, including ADF~\cite{tu2020rgbt}, MIDD~\cite{tu2021multi}, CSRNet~\cite{huo2021efficient}, CGFNet~\cite{wang2021cgfnet}, SwinNet~\cite{liu2021swinnet}, OSRNet~\cite{huo2022real}, TNet~\cite{TNet}, DCNet~\cite{tu2022weakly}, MCFNet~\cite{ma2023modal}, HRTransNet~\cite{Tang2023RTransNet}, LSNet~\cite{zhou2023lsnet}, CAVER~\cite{pang2023caver}, WaveNet~\cite{zhou2023wave}, and SPNet~\cite{zhang23SPNet}. For a fair comparison, we use the published code with the default parameters to implement these methods. For methods without the published source code, we directly use the results provided by the authors.

It is worth noting that our method does not need to remove the model module when handling alignment settings. There are several reasons for this. Firstly, in the case of alignment, objects in RGB and thermal modalities correspond in position and scale. Since the correlation modeling of the ACM module is performed in asymmetric window pairs of the two modalities, and the large window is obtained by expanding the small one, the corresponding information of the two modalities can still be completely covered. In addition, although the difference in window size introduces some interference information, the ACM is still able to establish sufficient multi-modal correlations in the aligned image pairs through the semantic guidance and powerful correlation operation. Secondly, being unaffected by unalignment, the AFSM module can calculate more accurate offsets to sample more relevant saliency features from the thermal modality through cascaded deformable convolutions.

\textit{1) Quantitative Evaluation.} 
Table~\ref{tab:compare} shows the quantitative comparison results. It can be seen that our method overall outperforms the compared methods on the aligned, weakly aligned, and unaligned datasets. Compared with the suboptimal method (i.e., SPNet~\cite{zhang23SPNet}), our proposed SACNet achieves average improvements of 2.5\%, 1.2\%, 3.2\%, and 24.2\% on the four evaluation metrics (i.e., ${E_\xi }$, ${S_\alpha }$, ${F_\beta ^\omega}$, and $MAE$) of the seven datasets, respectively. By replacing our backbone with CNN-based ResNet50, our method still has comparable performance, especially compared with methods~\cite{pang2023caver,ma2023modal,TNet} that also use ResNet50 as the backbone, which further demonstrates the effectiveness of our method. In addition, the computational complexity of our model is high, which mainly comes from the correlation operation and the cascaded deformable convolutions. Comparing the results on the aligned datasets and their corresponding weakly aligned datasets, we find that all compared methods suffer severe performance degradation on the weakly aligned datasets. This is mainly because they fail to establish robust correlations between the unaligned and weakly aligned modalities. In contrast, the performance of our method on the three weakly aligned datasets is closed to that on the corresponding aligned ones. This indicates that our method can better maintain the performance with a slight interference from misalignment. For the newly constructed UVT2000 dataset, the overall performance of all methods is inferior to that of the existing datasets, which indicates that alignment-free RGBT SOD is challenging and has great potential. 
Compared to the suboptimal method DCNet~\cite{tu2022weakly} on the challenging UVT2000 dataset, the minimum percentage gains on the four evaluation metrics (i.e., ${E_\xi }$, ${S_\alpha }$, ${F_\beta ^\omega}$, and $MAE$) of our method are 7.8\%, 10.4\%, 24.8\%, and 36.1\%, respectively. Furthermore, the PR curves of our method and the compared methods are presented in Fig.~\ref{fig::pr}. It can be seen that the curves of our method are overall more upward on the seven datasets, indicating that the saliency maps predicted by our method have a higher confidence and accuracy.

\begin{table*}[t]
	\centering
	\caption{Ablation Studies on One Unaligned, Three Weakly Aligned, and Three Aligned Datasets. The Best Results Are Marked with \textbf{Bold}.}
	\resizebox{1\textwidth}{!}{
		\begin{tabular}{c|ccc|ccc|ccc|ccc|ccc|ccc|ccc}
			\toprule
			\multirow{2}[4]{*}{models} & \multicolumn{3}{c|}{UVT2000} & \multicolumn{3}{c|}{un-VT5000} & \multicolumn{3}{c|}{un-VT1000} & \multicolumn{3}{c|}{un-VT821} & \multicolumn{3}{c|}{VT5000} & \multicolumn{3}{c|}{VT1000} & \multicolumn{3}{c}{VT821} \\
			\cmidrule{2-22}          & ${E_\xi }$ $\uparrow$     & ${F_\beta ^\omega}$ $\uparrow$    & $MAE$ \hfill$\downarrow$   & ${E_\xi }$ $\uparrow$     & ${F_\beta ^\omega}$ $\uparrow$    & $MAE$ $\downarrow$   & ${E_\xi }$ $\uparrow$     & ${F_\beta ^\omega}$ $\uparrow$    & $MAE$ $\downarrow$   & ${E_\xi }$ $\uparrow$     & ${F_\beta ^\omega}$ $\uparrow$    & $MAE$ $\downarrow$   & ${E_\xi }$ $\uparrow$    & ${F_\beta ^\omega}$ $\uparrow$    & $MAE$ $\downarrow$   & ${E_\xi }$ $\uparrow$    & ${F_\beta ^\omega}$ $\uparrow$    & $MAE$ $\downarrow$  & ${E_\xi }$ $\uparrow$    & ${F_\beta ^\omega}$ $\uparrow$    & $MAE$ $\downarrow$ \\
			\midrule
			SACNet & \textbf{0.812} & \textbf{0.640} & \textbf{0.036} & \textbf{0.949} & \textbf{0.876} & \textbf{0.023} & \textbf{0.954} & \textbf{0.923} & \textbf{0.014} & \textbf{0.929} & \textbf{0.857} & \textbf{0.026} & \textbf{0.957} & \textbf{0.888} & \textbf{0.021} & \textbf{0.958} & \textbf{0.927} & \textbf{0.014} & \textbf{0.932} & \textbf{0.859} & \textbf{0.025} \\
			$w/o$ ACM & 0.777  & 0.603  & 0.049  & 0.939  & 0.861  & 0.028  & 0.947  & 0.912  & 0.017  & 0.915  & 0.836  & 0.031  & 0.943  & 0.867  & 0.026  & 0.949  & 0.915  & 0.016  & 0.917  & 0.841  & 0.030  \\
			$w/o$ AWP & 0.779  & 0.607  & 0.049  & 0.942  & 0.865  & 0.027  & 0.947  & 0.916  & 0.016  & 0.917  & 0.839  & 0.030  & 0.950  & 0.878  & 0.023  & 0.951  & 0.919  & 0.015  & 0.923  & 0.846  & 0.028  \\
			$w/o$ SGM & 0.790  & 0.612  & 0.044  & 0.944  & 0.869  & 0.026  & 0.948  & 0.914  & 0.016  & 0.918  & 0.844  & 0.029  & 0.951  & 0.881  & 0.023  & 0.953  & 0.919  & 0.015  & 0.923  & 0.848  & 0.028  \\
			$w/o$ AFSM & 0.785  & 0.618  & 0.045  & 0.941  & 0.865  & 0.027  & 0.948  & 0.913  & 0.016  & 0.919  & 0.840  & 0.030  & 0.946  & 0.873  & 0.024  & 0.949  & 0.914  & 0.016  & 0.922  & 0.845  & 0.029  \\
			\bottomrule
		\end{tabular}%
	}
	\label{tab:ablation}%
\end{table*}%
\begin{figure}[t]
	\centering
	\includegraphics[width=1\columnwidth]{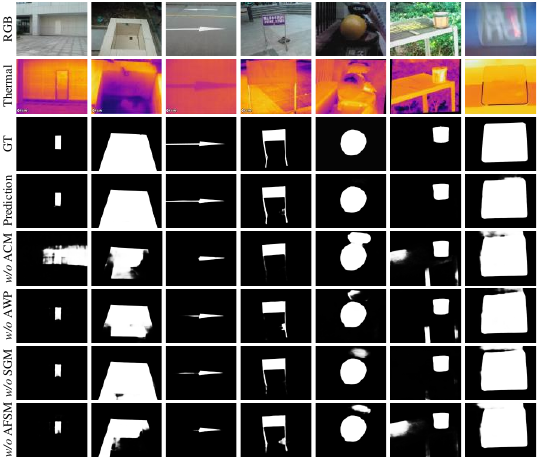}
	\caption{Visual ablation experiments for each component in different scenes. The first to fourth columns are unaligned samples, the fifth and sixth columns are weakly aligned samples, and the seventh column are aligned samples.}
	\label{fig::ablation_com}
\end{figure}
\begin{figure}[t]
	\centering
	\includegraphics[width=1\columnwidth]{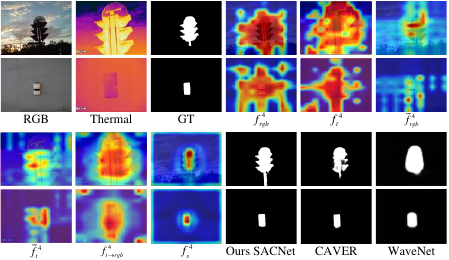}
	\caption{Feature visualization for ACM and AFSM in the highest layer of backbone. $f_s^i$ and $f_t^4$ are the extracted highest-layer features, $\tilde{f}_{rgb}^4$ and $\tilde{f}_{t}^4$ are features enhanced by ACM, $\tilde{f}_{t \to rgb}^4$ is the sampled thermal feature by AFSM, and $f_s^4$ is the integrated multi-modal feature. The comparison with advanced methods (i.e., CAVER~\cite{pang2023caver} and WaveNet~\cite{zhou2023wave}) demonstrates the effectiveness of our method.} 
	\label{fig::visual}
\end{figure}
\textit{2) Qualitative Evaluation.} 
Fig.~\ref{fig::dingxing} visually illustrates the qualitative comparison under various challenging scenes, including bad weather (e.g., Row 1), multiple salient objects (e.g., Row 2), center bias (e.g., Row 3), small salient
object (e.g., Row 4), similar foreground and background (e.g., Row 5), and low illumination (e.g., Row 6). For a comprehensive comparison, the image pairs in the first two rows are from our newly constructed UVT2000 dataset, the third and fourth rows are from the weakly aligned datasets, and the last two rows are from the manually aligned datasets. 
It can be seen that our method is able to predict the salient regions accurately compared to both the method (i.e., DCNet~\cite{tu2022weakly}) for weakly aligned image pairs and the transformer-based methods (i.e., SPNet~\cite{zhang23SPNet}, WaveNet~\cite{zhou2023wave}, CAVER~\cite{pang2023caver}, HRTransNet~\cite{Tang2023RTransNet}, and SwinNet~\cite{liu2021swinnet}). This shows that our method can establish strong correlations between both aligned and unaligned modalities and make full use of the multi-modal complementary information.

\subsection{Ablation Studies}
In this section, we perform ablation studies to illustrate the effectiveness of the components in our method. The results for unaligned, weakly aligned, and aligned datasets are shown in Table~\ref{tab:ablation}, in which the first line (i.e., SACNet) shows the performance of our full model. '$w/o$' means to disable the corresponding component. In addition, more detailed ablation experiments for asymmetric window partition (AWP) and associated feature sampling module (AFSM) are shown in Table~\ref{tab:ablation} and Table~\ref{tab:dconv}, respectively. We also present some ablation analyses about the visualization results in Fig.~\ref{fig::ablation_com} and Fig.~\ref{fig::visual}.

\textit{1) Effectiveness of ACM.} 
In order to verify the effectiveness of the proposed ACM, we directly remove it, denoted as '$w/o$ ACM' in Table~\ref{tab:ablation}. This means that the correlation of the two modalities cannot be modeled adequately, especially on the unaligned and weakly aligned datasets. Compared with our full model SACNet, the performance on $MAE$ metric drops by 36.1\%, 21.7\%, 21.4\%, and 19.2\% on UVT2000, un-VT5000, un-VT1000, and un-VT821 datasets, respectively. On the aligned VT821 dataset, the performance on the three evaluation metrics (${E_\xi }$, ${F_\beta ^\omega}$, and $MAE$) separately decreases by 1.6\%, 2.1\%, and 20.0\%. The corresponding visual results in the fifth row of Fig.~\ref{fig::ablation_com} prove that without the correlation modeling of ACM, the model will focus on irrelevant information and introduce some noise. Fig.~\ref{fig::visual} further demonstrates the visualization of RGB and thermal features (i.e., $f_{rgb}^4$ and $\tilde{f}_{rgb}^4$, ${f}_{t}^4$ and $\tilde{f}_{t}^4$) before and after the ACM. With the correlation modeling of ACM, both the RGB and thermal features focus more on the objects, alleviating the interference of misalignment. This confirms that the ACM can effectively model multi-modal correlations and exploit their complementarity for salient regions.

\begin{table}[t]
	\centering
	\caption{Ablation Studies on Small Window Size $M$ and Large Window Size $N$ of Asymmetric Window Pairs. The Best Results Are Marked with \textbf{Bold}.}
	\resizebox{1\columnwidth}{!}{
		\begin{tabular}{c|ccc|ccc|ccc}
			\toprule
			\multicolumn{1}{c|}{\multirow{2}[4]{*}{Models}} & \multicolumn{3}{c|}{UVT2000} & \multicolumn{3}{c|}{un-VT821} & \multicolumn{3}{c}{VT821} \\
			\cmidrule{2-10}          & ${E_\xi }$ \hfill$\uparrow$     & ${F_\beta ^\omega}$ \hfill$\uparrow$    & $MAE$ \hfill$\downarrow$   & ${E_\xi }$ \hfill$\uparrow$     & ${F_\beta ^\omega}$ \hfill$\uparrow$    & $MAE$ \hfill$\downarrow$   & ${E_\xi }$ \hfill$\uparrow$    & ${F_\beta ^\omega}$ \hfill$\uparrow$    & $MAE$ \hfill$\downarrow$ \\
			\midrule
			\multicolumn{1}{c|}{SACNet ($M$=4, $N$=6)} & \textbf{0.812} & \textbf{0.640} & \textbf{0.036} & \textbf{0.929} & \textbf{0.857} & \textbf{0.026} & \textbf{0.932} & \textbf{0.859} & \textbf{0.025} \\
			$M$=4, $N$=4 & 0.789  & 0.621  & 0.040  & 0.920  & 0.845  & 0.029  & 0.928  & 0.852  & 0.028  \\
			$M$=6, $N$=6 & 0.789  & 0.613  & 0.044  & 0.920  & 0.844  & 0.029  & 0.928  & 0.853  & 0.026  \\
			$M$=3, $N$=4 & 0.798  & 0.625  & 0.039  & 0.923  & 0.850  & 0.028  & 0.928  & 0.853  & 0.026  \\
			$M$=2, $N$=4 & 0.801  & 0.629  & 0.038  & 0.923  & 0.841  & 0.029  & 0.930  & 0.841  & 0.028  \\
			$M$=2, $N$=6 & 0.773  & 0.601  & 0.050  & 0.914  & 0.837  & 0.031  & 0.919  & 0.842  & 0.031  \\
			\bottomrule
		\end{tabular}%
	}
	\label{tab:awp}%
\end{table}%

\begin{table}[t]
	\centering
	\caption{Ablation Studies on the Number of Cascaded Deformable Convolutions $N$. The Best Results Are Marked with \textbf{Bold}.}
	\resizebox{1\columnwidth}{!}{
		\begin{tabular}{c|ccc|ccc|ccc}
			\toprule
			\multirow{2}[4]{*}{Models} & \multicolumn{3}{c|}{UVT2000} & \multicolumn{3}{c|}{un-VT821} & \multicolumn{3}{c}{VT821} \\
			\cmidrule{2-10}          & ${E_\xi }$ \hfill$\uparrow$     & ${F_\beta ^\omega}$ \hfill$\uparrow$    & $MAE$ \hfill$\downarrow$   & ${E_\xi }$ \hfill$\uparrow$     & ${F_\beta ^\omega}$ \hfill$\uparrow$    & $MAE$ \hfill$\downarrow$   & ${E_\xi }$ \hfill$\uparrow$    & ${F_\beta ^\omega}$ \hfill$\uparrow$    & $MAE$ \hfill$\downarrow$ \\
			\midrule
			SACNet ($N$=4) & \textbf{0.812} & \textbf{0.640} & \textbf{0.036} & \textbf{0.929} & \textbf{0.857} & \textbf{0.026} & \textbf{0.932} & \textbf{0.859} & \textbf{0.025} \\
			$N$=1   & 0.793  & 0.621  & 0.043  & 0.921  & 0.847  & 0.029  & 0.923  & 0.849  & 0.027  \\
			$N$=2   & 0.795  & 0.621  & 0.042  & 0.924  & 0.850  & 0.030  & 0.926  & 0.852  & 0.027  \\
			$N$=3   & 0.794  & 0.625  & 0.040  & 0.926  & 0.853  & 0.029  & 0.929  & 0.854  & 0.026  \\
			$N$=5   & 0.802  & 0.631  & 0.040  & 0.927  & 0.856  & 0.028  & 0.929  & 0.857  & \textbf{0.025} \\
			\bottomrule
		\end{tabular}%
	}
	\label{tab:dconv}%
\end{table}%
\textit{2) Effectiveness of AWP.} 
We also directly disable the asymmetric window pairs, which means that the correlation modeling is performed within two whole feature maps. The results in the third row (i.e., $w/o$ AWP) of Table~\ref{tab:ablation} and the sixth row of Fig.~\ref{fig::ablation_com} demonstrate the positive effect of the AWP strategy. In particular, on the unaligned and weakly aligned datasets, without AWP, the $MAE$ metrics decreased by an average of 20.8\%. Note that the AWP is also valid for aligned datasets. The reason may be that under small window differences, irrelevant information is not introduced too much, and richer feature representations can be learned through the asymmetric window attention to facilitate the identification of salient regions. 

The window size of asymmetric window pairs is an important hyper-parameter of the proposed asymmetric window partition (AWP). Therefore, we set five pairs of values for the size of small window $M$ and large window $N$ as candidate values, and record the performance of the five variants on both aligned and unaligned datasets. The results are shown in Table~\ref{tab:awp}, in which the first line (i.e., SACNet) shows the performance of our complete model with a small window of size 4 (i.e., $M=4$) and a large window of size 6 (i.e., $N=6$). We first replace the proposed asymmetric window pairs with symmetric ones. The results are shown in the second (i.e., $M=4$, $N=4$) and third (i.e., $M=6$, $N=6$) rows of Table~\ref{tab:awp}. Compared with the complete model, we can find that the asymmetric window pair works on both aligned and unaligned datasets. 
In addition, we also conduct experiments on the special case with a large window size difference (i.e., $M=2$, $N=6$), and the result is shown in the last row of Table~\ref{tab:awp}. It can be seen that the large size difference between window pairs can cause performance degradation. This is mainly because the large window size difference introduces too much irrelevant information and background noise, disturbing the correlation modeling of salient regions.

\textit{3) Effectiveness of SGM.} 
To observe the effectiveness of the SGM, we remove it and perform the correlation operation directly on the original asymmetric window pairs. The results are shown in the forth row (i.e., $w/o$ SGM) of Table~\ref{tab:ablation} and. By comparing with the full model, it can be found that the SGM improves the three indicators on all seven datasets, especially obtains an average gain of 14.3\% on the $MAE$ metric for the unaligned and weakly aligned datasets. 
The visual results in the seventh row of Fig.~\ref{fig::ablation_com} show that without SGM, the model is able to segment rough but inaccurate salient regions. This suggests that semantic information indeed guides the correlation modeling of the ACM to focus on salient regions, thereby reducing missed and false detection.

\begin{table*}[t]
	\centering
	\caption{Quantitative Comparison of E-measure (${E_\xi }$), S-measure (${S_\alpha }$), Weighed F-measure (${F_\beta ^\omega}$), and Mean Absolute Error ($MAE$) on Six RGBD Datasets. The Best Three Results Are Marked with \textcolor[rgb]{ .792,  0,  0}{\textbf{Red}}, \textcolor[rgb]{ 0,  .533,  .2}{\textbf{Green}}, and \textcolor[rgb]{ 0,  .4,  .8}{\textbf{Blue}}.}
	\resizebox{1\textwidth}{!}{
		\begin{tabular}{cl|ccccccccccccccc|cc}
			\toprule
			\multicolumn{2}{c|}{\multirow{2}[2]{*}{Method}} & CCAFNet$_{21}$ & CDNet$_{21}$ & HAINet$_{21}$ & DFM$_{21}$ & SPNet$_{21}$ & RD3D$_{21}$ & DSA2F$_{21}$ & DCF$_{21}$ & VST$_{21}$ & MobileSal$_{22}$ & SSL$_{22}$ & DIGRNet$_{22}$ & LSNet$_{23}$ & CAVER$_{23}$ & PICRNet$_{23}$ & SACNet & SACNet \\
			\multicolumn{2}{c|}{} &\cite{zhou2021ccafnet}       &\cite{jin2021cdnet}       &\cite{li2021hierarchical}       &\cite{zhang2021depth}       &\cite{zhouiccv2021}       &\cite{chen2021rgb}       &\cite{Sun2021DeepRS}       &\cite{Ji_2021_DCF}       &\cite{liu2021visual}       &\cite{wu2021mobilesal}       &\cite{zhao2022self}       &\cite{cheng2022depth}       &\cite{zhou2023lsnet}       &\cite{pang2023caver}       &\cite{cong23picrnet}       & Ours  & Ours \\
			\midrule
			\multicolumn{2}{c|}{Backbone} & VGG16 & VGG16 & VGG16 & MobileNet-v2 & Res2Net50 & I3DResNet & VGG19 & ResNet50 & T2T-ViT & MobileNet-v2 & VGG16 & ResNet50 & MobileNet-v2 & ResNet50 & SwinT & ResNet50 & SwinB \\
			\midrule
			\multicolumn{2}{c|}{FPS \hfill$\uparrow$} & 88    & 86    & 11    & \textcolor[rgb]{ 0,  .4,  .8}{\textbf{252}} & 50    & 94    & 24    & 57    & 69    & \textcolor[rgb]{ 0,  .533,  .2}{\textbf{268}} & 52    & 33    & \textcolor[rgb]{ .792,  0,  0}{\textbf{316}} & 67    & 63    & 19    & 27 \\
			\multicolumn{2}{c|}{Parameters(M) \hfill$\downarrow$} & 41.8  & 32.4  & 59.8  & \textcolor[rgb]{ .792,  0,  0}{\textbf{2.2}} & 150.3  & 28.9  & 34.0  & 97.0  & 53.5  & \textcolor[rgb]{ 0,  .4,  .8}{\textbf{6.5}} & 74.2  & 166.7  & \textcolor[rgb]{ 0,  .533,  .2}{\textbf{4.6}} & 55.8  & 86.0  & 530.9  & 327.7  \\
			\midrule
			\multirow{4}[2]{*}{\begin{sideways}DUT\end{sideways}} & ${E_\xi}$ ~\hfill\hfill$\uparrow$   & 0.940  & 0.936  & 0.937  & 0.898  & 0.876  & 0.949  & 0.950  & 0.952  & \textcolor[rgb]{ 0,  .533,  .2}{\textbf{0.960}} & 0.936  & 0.927  & 0.948  & 0.927  & \textcolor[rgb]{ 0,  .4,  .8}{\textbf{0.955}} & \textcolor[rgb]{ .792,  0,  0}{\textbf{0.967}} & 0.953  & \textcolor[rgb]{ .792,  0,  0}{\textbf{0.967}} \\
			& ${S_\alpha}$ ~\hfill\hfill$\uparrow$   & 0.904  & 0.905  & 0.909  & 0.856  & 0.803  & \textcolor[rgb]{ 0,  .4,  .8}{\textbf{0.932}} & 0.921  & 0.924  & \textcolor[rgb]{ 0,  .533,  .2}{\textbf{0.943}} & 0.896  & 0.889  & 0.926  & 0.886  & 0.931  & \textcolor[rgb]{ 0,  .533,  .2}{\textbf{0.943}} & 0.923  & \textcolor[rgb]{ .792,  0,  0}{\textbf{0.946}} \\
			& ${F_\beta ^\omega}$ ~\hfill\hfill$\uparrow$   & 0.884  & 0.878  & 0.887  & 0.795  & 0.747  & 0.913  & 0.914  & 0.913  & 0.926  & 0.869  & 0.859  & 0.902  & 0.856  & \textcolor[rgb]{ 0,  .4,  .8}{\textbf{0.920}} & \textcolor[rgb]{ .792,  0,  0}{\textbf{0.935}} & 0.912  & \textcolor[rgb]{ .792,  0,  0}{\textbf{0.944}} \\
			& $MAE$\hfill$\downarrow$  & 0.037  & 0.039  & 0.038  & 0.062  & 0.085  & 0.031  & 0.030  & 0.030  & \textcolor[rgb]{ 0,  .4,  .8}{\textbf{0.025}} & 0.044  & 0.046  & 0.033  & 0.049  & 0.029  & \textcolor[rgb]{ 0,  .533,  .2}{\textbf{0.022}} & 0.030  & \textcolor[rgb]{ .792,  0,  0}{\textbf{0.021}} \\
			\midrule
			\multirow{4}[2]{*}{\begin{sideways}NJUD\end{sideways}} & ${E_\xi}$ ~\hfill\hfill$\uparrow$   & 0.920  & 0.903  & 0.917  & 0.913  & \textcolor[rgb]{ 0,  .533,  .2}{\textbf{0.931}} & 0.918  & 0.923  & 0.922  & 0.913  & 0.914  & 0.881  & 0.928  & 0.891  & 0.922  & \textcolor[rgb]{ 0,  .4,  .8}{\textbf{0.930}} & \textcolor[rgb]{ 0,  .4,  .8}{\textbf{0.930}} & \textcolor[rgb]{ .792,  0,  0}{\textbf{0.941}} \\
			& ${S_\alpha}$ ~\hfill\hfill$\uparrow$   & 0.910  & 0.872  & 0.909  & 0.906  & \textcolor[rgb]{ 0,  .4,  .8}{\textbf{0.925}} & 0.915  & 0.903  & 0.903  & 0.922  & 0.905  & 0.841  & \textcolor[rgb]{ 0,  .533,  .2}{\textbf{0.932}} & 0.837  & 0.920  & 0.924  & 0.921  & \textcolor[rgb]{ .792,  0,  0}{\textbf{0.935}} \\
			& ${F_\beta ^\omega}$ ~\hfill\hfill$\uparrow$   & 0.883  & 0.828  & 0.882  & 0.868  & \textcolor[rgb]{ 0,  .4,  .8}{\textbf{0.909}} & 0.890  & 0.889  & 0.884  & 0.892  & 0.874  & 0.786  & \textcolor[rgb]{ 0,  .4,  .8}{\textbf{0.909}} & 0.775  & 0.903  & \textcolor[rgb]{ 0,  .4,  .8}{\textbf{0.909}} & \textcolor[rgb]{ 0,  .533,  .2}{\textbf{0.910}} & \textcolor[rgb]{ .792,  0,  0}{\textbf{0.933}} \\
			& $MAE$\hfill$\downarrow$  & 0.037  & 0.054  & 0.039  & 0.042  & \textcolor[rgb]{ 0,  .4,  .8}{\textbf{0.029}} & 0.037  & 0.039  & 0.038  & 0.035  & 0.040  & 0.065  & \textcolor[rgb]{ 0,  .533,  .2}{\textbf{0.028}} & 0.074  & 0.032  & 0.030  & \textcolor[rgb]{ 0,  .4,  .8}{\textbf{0.029}} & \textcolor[rgb]{ .792,  0,  0}{\textbf{0.022}} \\
			\midrule
			\multirow{4}[2]{*}{\begin{sideways}NLPR\end{sideways}} & ${E_\xi}$ ~\hfill\hfill$\uparrow$   & 0.951  & 0.951  & 0.951  & 0.945  & 0.957  & 0.957  & 0.950  & 0.956  & 0.953  & 0.950  & 0.954  & 0.955  & 0.955  & \textcolor[rgb]{ 0,  .4,  .8}{\textbf{0.959}} & \textcolor[rgb]{ .792,  0,  0}{\textbf{0.965}} & 0.951  & \textcolor[rgb]{ 0,  .533,  .2}{\textbf{0.964}} \\
			& ${S_\alpha}$ ~\hfill\hfill$\uparrow$   & 0.922  & 0.925  & 0.921  & 0.923  & 0.928  & 0.929  & 0.918  & 0.921  & \textcolor[rgb]{ 0,  .4,  .8}{\textbf{0.931}} & 0.920  & 0.919  & \textcolor[rgb]{ .792,  0,  0}{\textbf{0.935}} & 0.918  & 0.929  & \textcolor[rgb]{ .792,  0,  0}{\textbf{0.935}} & 0.920  & \textcolor[rgb]{ .792,  0,  0}{\textbf{0.935}} \\
			& ${F_\beta ^\omega}$ ~\hfill\hfill$\uparrow$   & 0.883  & 0.886  & 0.884  & 0.876  & \textcolor[rgb]{ 0,  .4,  .8}{\textbf{0.899}} & 0.894  & 0.889  & 0.892  & 0.891  & 0.878  & 0.885  & 0.895  & 0.881  & \textcolor[rgb]{ 0,  .4,  .8}{\textbf{0.899}} & \textcolor[rgb]{ 0,  .533,  .2}{\textbf{0.911}} & 0.888  & \textcolor[rgb]{ .792,  0,  0}{\textbf{0.917}} \\
			& $MAE$\hfill$\downarrow$  & 0.026  & 0.025  & 0.025  & 0.026  & \textcolor[rgb]{ 0,  .533,  .2}{\textbf{0.021}} & \textcolor[rgb]{ 0,  .4,  .8}{\textbf{0.022}} & 0.024  & 0.023  & 0.024  & 0.025  & 0.027  & 0.023  & 0.024  & \textcolor[rgb]{ 0,  .4,  .8}{\textbf{0.022}} & \textcolor[rgb]{ .792,  0,  0}{\textbf{0.019}} & 0.024  & \textcolor[rgb]{ .792,  0,  0}{\textbf{0.019}} \\
			\midrule
			\multirow{4}[2]{*}{\begin{sideways}SSD\end{sideways}} & ${E_\xi}$ ~\hfill\hfill$\uparrow$   & \textcolor[rgb]{ 0,  .533,  .2}{\textbf{0.915}} & 0.849  & 0.843  & 0.871  & \textcolor[rgb]{ 0,  .4,  .8}{\textbf{0.910}} & 0.905  & 0.904  & 0.898  & 0.907  & 0.898  & 0.833  & 0.889  & 0.902  & \textcolor[rgb]{ 0,  .533,  .2}{\textbf{0.915}} & \textcolor[rgb]{ 0,  .533,  .2}{\textbf{0.915}} & \textcolor[rgb]{ 0,  .533,  .2}{\textbf{0.915}} & \textcolor[rgb]{ .792,  0,  0}{\textbf{0.929}} \\
			& ${S_\alpha}$ ~\hfill\hfill$\uparrow$   & 0.876  & 0.798  & 0.769  & 0.814  & 0.871  & 0.863  & 0.876  & 0.852  & \textcolor[rgb]{ 0,  .533,  .2}{\textbf{0.889}} & 0.863  & 0.745  & 0.866  & 0.856  & 0.874  & \textcolor[rgb]{ 0,  .4,  .8}{\textbf{0.878}} & 0.876  & \textcolor[rgb]{ .792,  0,  0}{\textbf{0.896}} \\
			& ${F_\beta ^\omega}$ ~\hfill\hfill$\uparrow$   & \textcolor[rgb]{ 0,  .533,  .2}{\textbf{0.839}} & 0.706  & 0.682  & 0.733  & 0.831  & 0.794  & 0.836  & 0.800  & 0.836  & 0.804  & 0.638  & 0.804  & 0.796  & 0.826  & \textcolor[rgb]{ 0,  .4,  .8}{\textbf{0.837}} & 0.835  & \textcolor[rgb]{ .792,  0,  0}{\textbf{0.870}} \\
			& $MAE$\hfill$\downarrow$ & 0.054  & 0.073  & 0.101  & 0.076  & \textcolor[rgb]{ 0,  .4,  .8}{\textbf{0.044}} & 0.052  & 0.047  & 0.053  & 0.045  & 0.052  & 0.100  & 0.053  & 0.055  & \textcolor[rgb]{ 0,  .4,  .8}{\textbf{0.044}} & 0.046  & \textcolor[rgb]{ 0,  .533,  .2}{\textbf{0.042}} & \textcolor[rgb]{ .792,  0,  0}{\textbf{0.032}} \\
			\midrule
			\multirow{4}[2]{*}{\begin{sideways}SIP\end{sideways}} & ${E_\xi}$ ~\hfill\hfill$\uparrow$   & 0.915  & 0.913  & 0.924  & 0.919  & 0.930  & 0.919  & 0.908  & 0.920  & \textcolor[rgb]{ .792,  0,  0}{\textbf{0.936}} & 0.914  & 0.921  & 0.918  & 0.911  & 0.927  & 0.916  & \textcolor[rgb]{ 0,  .4,  .8}{\textbf{0.932}} & \textcolor[rgb]{ 0,  .533,  .2}{\textbf{0.934}} \\
			& ${S_\alpha}$ ~\hfill\hfill$\uparrow$   & 0.876  & 0.872  & 0.886  & 0.883  & 0.894  & 0.885  & 0.861  & 0.873  & \textcolor[rgb]{ 0,  .533,  .2}{\textbf{0.903}} & 0.873  & 0.880  & 0.885  & \textcolor[rgb]{ .792,  0,  0}{\textbf{0.909}} & 0.893  & 0.865  & 0.888  & \textcolor[rgb]{ 0,  .4,  .8}{\textbf{0.896}} \\
			& ${F_\beta ^\omega}$ ~\hfill\hfill$\uparrow$   & 0.839  & 0.839  & 0.860  & 0.844  & 0.873  & 0.852  & 0.838  & 0.850  & \textcolor[rgb]{ 0,  .533,  .2}{\textbf{0.878}} & 0.837  & 0.851  & 0.849  & \textcolor[rgb]{ 0,  .4,  .8}{\textbf{0.877}} & 0.874  & 0.838  & 0.871  & \textcolor[rgb]{ .792,  0,  0}{\textbf{0.889}} \\
			& $MAE$\hfill$\downarrow$  & 0.054  & 0.056  & 0.049  & 0.051  & \textcolor[rgb]{ 0,  .4,  .8}{\textbf{0.043}} & 0.049  & 0.057  & 0.051  & \textcolor[rgb]{ 0,  .533,  .2}{\textbf{0.040}} & 0.054  & 0.049  & 0.053  & \textcolor[rgb]{ 0,  .533,  .2}{\textbf{0.040}} & \textcolor[rgb]{ 0,  .4,  .8}{\textbf{0.043}} & 0.056  & 0.044  & \textcolor[rgb]{ .792,  0,  0}{\textbf{0.039}} \\
			\midrule
			\multirow{4}[2]{*}{\begin{sideways}STERE\end{sideways}} & ${E_\xi}$ ~\hfill\hfill$\uparrow$   & 0.921  & 0.929  & \textcolor[rgb]{ 0,  .4,  .8}{\textbf{0.930}} & 0.912  & \textcolor[rgb]{ 0,  .4,  .8}{\textbf{0.930}} & 0.926  & 0.928  & \textcolor[rgb]{ 0,  .533,  .2}{\textbf{0.931}} & 0.916  & 0.916  & 0.923  & 0.927  & 0.913  & \textcolor[rgb]{ 0,  .533,  .2}{\textbf{0.931}} & \textcolor[rgb]{ .792,  0,  0}{\textbf{0.937}} & \textcolor[rgb]{ 0,  .4,  .8}{\textbf{0.930}} & 0.929  \\
			& ${S_\alpha}$ ~\hfill\hfill$\uparrow$   & 0.891  & 0.907  & 0.909  & 0.898  & 0.907  & 0.911  & 0.897  & 0.905  & 0.913  & 0.903  & 0.897  & \textcolor[rgb]{ 0,  .4,  .8}{\textbf{0.916}} & 0.871  & 0.914  & \textcolor[rgb]{ .792,  0,  0}{\textbf{0.920}} & 0.902  & \textcolor[rgb]{ 0,  .533,  .2}{\textbf{0.917}} \\
			& ${F_\beta ^\omega}$ ~\hfill\hfill$\uparrow$   & 0.853  & 0.871  & 0.877  & 0.850  & 0.879  & 0.877  & 0.877 & 0.880  & 0.872 & 0.865  & 0.864  & 0.877  & 0.827  & \textcolor[rgb]{ 0,  .4,  .8}{\textbf{0.887}} & \textcolor[rgb]{ 0,  .533,  .2}{\textbf{0.898}} & 0.879  & \textcolor[rgb]{ .792,  0,  0}{\textbf{0.901}} \\
			& $MAE$\hfill$\downarrow$  & 0.044  & 0.039  & 0.038  & 0.045  & 0.037  & 0.038  & 0.038  & 0.037  & 0.038  & 0.041  & 0.042  & 0.038  & 0.054  & \textcolor[rgb]{ 0,  .4,  .8}{\textbf{0.034}} & \textcolor[rgb]{ 0,  .533,  .2}{\textbf{0.031}} & 0.038  & \textcolor[rgb]{ .792,  0,  0}{\textbf{0.030}} \\
			\bottomrule
		\end{tabular}%
	}
	\label{tab:compare_rgbd}%
\end{table*}%

\textit{4) Effectiveness of AFSM.}
We also replace the AFSM with the concatenation and convolution operations, which means that the spatially inconsistent features of the two modalities are directly integrated. The results in the last row (i.e., $w/o$ AFSM) of Table~\ref{tab:ablation} and Fig.~\ref{fig::ablation_com} demonstrate the effectiveness of AFSM. On all seven datasets, without AFSM, the three evaluation metrics (i.e., ${E_\xi }$, ${F_\beta ^\omega}$, and $MAE$) drop by an average of 1.4\%, 1.9\%, and 15.4\%, respectively. As shown in Fig.~\ref{fig::visual}, the thermal feature (i.e., $\tilde{f}_{t \to rgb}^4$) sampled according to the RGB feature captures the salient regions completely. Based on this, the integrated multi-modal feature (i.e., $f_{s}^4$) can locate the salient regions accurately.

We also complement the ablation experiments on the number of cascaded deformable convolutions in AFSM, with the results shown in Table~\ref{tab:dconv}. As the number of deformable convolutions increases, AFSM can sample more relevant multi-modal features for accurate integration, with improved performance. However, when the number of deformable convolutions increases to 5, the model shows a decreasing trend. This is mainly because too many deformable convolutions introduce a large number of parameters, leading to overfitting of the model.

\subsection{Experiment on RGBD SOD Datasets}
RGBD salient object detection is another multi-modal SOD task, which utilizes the complementary information of RGB images and depth maps. To further verify the effectiveness of our method, we perform experiments on six RGBD SOD datasets, including DUT-RGBD (i.e., DUT)~\cite{piao2019depth}, NJUD~\cite{ju2014depth}, NLPR~\cite{peng2014benchmark}, SIP~\cite{fan2020rethinking}, SSD~\cite{zhu2017three} and STERE~\cite{niu2012leveraging}. 

\textit{1) Datasets:}
NJUD contains 1985 pairs of visible images and depth maps that are collected from the Internet, 3D movies, and stereo photos. 
DUT-RGBD contains 1200 pairs of images captured from both indoor and outdoor scenes, which include a large number of image clutter challenges. NJUD and NLPR consist of 1985 and 1000 pairs of images separately, which contain more scenarios with the challenge of scale variation. SIP incorporates 929 paired images about persons from different angles. SSD comprises 80 pairs of RGBD images from several movies. STERE collects 1000 pairs of RGBD	images from the Internet.
Following~\cite{Sun2021DeepRS}, we take a collection of 700 samples from NLPR, 1485 samples from NJUD, and 800 samples from DUT-RGBD as our RGBD training set. 

\textit{2) Experiment Setup:}
We compare our method with 15 state-of-the-art RGBD SOD methods, including CCAFNet~\cite{zhou2021ccafnet}, CDNet~\cite{jin2021cdnet}, HAINet~\cite{li2021hierarchical}, DFM~\cite{zhang2021depth}, SPNet~\cite{zhouiccv2021}, RD3D~\cite{chen2021rgb}, DSA2F~\cite{Sun2021DeepRS}, DCF~\cite{Ji_2021_DCF}, VST~\cite{liu2021visual}, MobileSal~\cite{wu2021mobilesal}, SSL~\cite{zhao2022self}, DIGRNet~\cite{cheng2022depth}, LSNet~\cite{zhou2023lsnet}, CAVER~\cite{pang2023caver}, and PICRNet~\cite{cong23picrnet}. All predicted results and code used in the experiments are released by the authors.

\textit{3) Quantitative Evaluation:}
Table~\ref{tab:compare_rgbd} shows the quantitative comparison results in terms of the four evaluation metrics (i.e., ${E_\xi }$, ${S_\alpha}$, ${F_\beta ^\omega}$, and $MAE$). It can be seen that our network outperforms all compared methods on the six datasets, except for ${E_\xi}$ and ${S_\alpha }$ on the SIP~\cite{fan2020rethinking} and STERE~\cite{niu2012leveraging} datasets, and ${E_\xi}$ on the NLPR dataset. Compared with the suboptimal method (i.e., PICRNet), our method achieve average improvements of the six datasets on the four evaluation metrics (${E_\xi }$, ${S_\alpha }$, ${F_\beta ^\omega}$, and $MAE$) by 0.9\%, 1.2\%, 2.4\%, and 22.0\%, respectively. By replacing the backbone with ResNet50, our method is still comparable to the advanced method CAVER, which also uses ResNet50 as the backbone.
This demonstrates that our method is also able to model the strong correlation between RGB and depth modalities and fully exploit their complementary information for saliency prediction.

\subsection{Experiment on single-modal SOD Datasets}
\begin{table}[t]
	\centering
	\caption{Quantitative Comparison of E-measure (${E_\xi }$), S-measure (${S_\alpha }$), Weighed F-measure (${F_\beta ^\omega}$), and Mean Absolute Error ($MAE$) on Five Single-modal Datasets. The Best Three Results Are Marked with \textcolor[rgb]{ .792,  0,  0}{\textbf{Red}}, \textcolor[rgb]{ 0,  .533,  .2}{\textbf{Green}}, and \textcolor[rgb]{ 0,  .4,  .8}{\textbf{Blue}}.}
	\resizebox{1\columnwidth}{!}{
		\begin{tabular}{cc|cccccc|c}
			\hline
			\multicolumn{2}{c|}{\multirow{2}[2]{*}{Method}} & VST$_{21}$ & ICON$_{22}$ & EDN$_{22}$ & MENet$_{23}$ & BBRF$_{23}$ & SelfReformer$_{23}$ & SACNet \bigstrut[t]\\
			\multicolumn{2}{c|}{} &\cite{liu2021visual}      &\cite{zhuge2022salient}       &\cite{wu2022edn}       &\cite{wang2023pixels}       &\cite{ma2023bbrf}       &\cite{yun2023towards}       & Ours \bigstrut[b]\\
			\hline
			\multicolumn{2}{c|}{Backbone} & T2T-ViT & SwinB & ResNet50 & ResNet50 & SwinB & PVT-v2 & SwinB \bigstrut\\
			\hline
			\multicolumn{2}{c|}{FPS $\uparrow$} & \textcolor[rgb]{ 0,  .533,  .2}{\textbf{100}} & 57    & \textcolor[rgb]{ .792,  0,  0}{\textbf{123}} & -     & \textcolor[rgb]{ 0,  .4,  .8}{\textbf{87}} & 69    & 27 \bigstrut[t]\\
			\multicolumn{2}{c|}{Parameters(M) $\downarrow$} & \textcolor[rgb]{ .792,  0,  0}{\textbf{32.2}} & 92.4  & \textcolor[rgb]{ 0,  .533,  .2}{\textbf{42.8}} & -     & 74.1  & \textcolor[rgb]{ 0,  .4,  .8}{\textbf{44.6}} & 327.7 \bigstrut[b]\\
			\hline
			\multirow{4}[2]{*}{\begin{sideways}DUTS\end{sideways}} & ${E_\xi}$ ~\hfill\hfill$\uparrow$   & 0.892  & \textcolor[rgb]{ 0,  .533,  .2}{\textbf{0.930}} & 0.908  & 0.921  & \textcolor[rgb]{ 0,  .4,  .8}{\textbf{0.927}} & 0.921  & \textcolor[rgb]{ .792,  0,  0}{\textbf{0.933}} \bigstrut[t]\\
			& ${S_\alpha}$ ~\hfill\hfill$\uparrow$   & 0.896  & \textcolor[rgb]{ 0,  .533,  .2}{\textbf{0.917}} & 0.892  & 0.905  & 0.909  & \textcolor[rgb]{ 0,  .4,  .8}{\textbf{0.911}} & \textcolor[rgb]{ .792,  0,  0}{\textbf{0.920}} \\
			& ${F_\beta ^\omega}$ ~\hfill\hfill$\uparrow$   & 0.828  & \textcolor[rgb]{ 0,  .533,  .2}{\textbf{0.886}} & 0.845  & 0.870  & \textcolor[rgb]{ 0,  .533,  .2}{\textbf{0.886}} & \textcolor[rgb]{ 0,  .4,  .8}{\textbf{0.872}} & \textcolor[rgb]{ .792,  0,  0}{\textbf{0.899}} \\
			& $MAE$\hfill$\downarrow$  & 0.037  & \textcolor[rgb]{ 0,  .533,  .2}{\textbf{0.025}} & 0.035  & 0.028  & \textcolor[rgb]{ 0,  .533,  .2}{\textbf{0.025}} & \textcolor[rgb]{ 0,  .4,  .8}{\textbf{0.027}} & \textcolor[rgb]{ .792,  0,  0}{\textbf{0.023}} \bigstrut[b]\\
			\hline
			\multirow{4}[1]{*}{\begin{sideways}ECSSD\end{sideways}} & ${E_\xi}$ ~\hfill\hfill$\uparrow$   & 0.918  & \textcolor[rgb]{ 0,  .4,  .8}{\textbf{0.932}} & 0.929  & 0.925  & \textcolor[rgb]{ 0,  .533,  .2}{\textbf{0.934}} & 0.929  & \textcolor[rgb]{ .792,  0,  0}{\textbf{0.937}} \bigstrut[t]\\
			& ${S_\alpha}$ ~\hfill\hfill$\uparrow$   & 0.932  & \textcolor[rgb]{ 0,  .533,  .2}{\textbf{0.941}} & 0.927  & 0.928  & \textcolor[rgb]{ 0,  .4,  .8}{\textbf{0.939}} & 0.936  & \textcolor[rgb]{ .792,  0,  0}{\textbf{0.948}} \\
			& ${F_\beta ^\omega}$ ~\hfill\hfill$\uparrow$   & 0.910  & \textcolor[rgb]{ 0,  .4,  .8}{\textbf{0.936}} & 0.918  & 0.920  & \textcolor[rgb]{ 0,  .533,  .2}{\textbf{0.944}} & 0.926  & \textcolor[rgb]{ .792,  0,  0}{\textbf{0.951}} \\
			& $MAE$\hfill$\downarrow$  & 0.033  & \textcolor[rgb]{ 0,  .4,  .8}{\textbf{0.023}} & 0.032  & 0.031  & \textcolor[rgb]{ 0,  .533,  .2}{\textbf{0.022}} & 0.027  & \textcolor[rgb]{ .792,  0,  0}{\textbf{0.019}}
			\bigstrut[b]\\
			\hline
			\multirow{4}[1]{*}{\begin{sideways}OMRON\end{sideways}} & ${E_\xi}$ ~\hfill\hfill$\uparrow$   & 0.861  & \textcolor[rgb]{ 0,  .533,  .2}{\textbf{0.898}} & 0.879  & 0.882  & \textcolor[rgb]{ 0,  .4,  .8}{\textbf{0.891}} & 0.889  & \textcolor[rgb]{ .792,  0,  0}{\textbf{0.904}} \\
			& ${S_\alpha}$ ~\hfill\hfill$\uparrow$   & 0.850  & \textcolor[rgb]{ 0,  .533,  .2}{\textbf{0.869}} & 0.849  & 0.850  & \textcolor[rgb]{ 0,  .4,  .8}{\textbf{0.861}} & \textcolor[rgb]{ 0,  .4,  .8}{\textbf{0.861}} & \textcolor[rgb]{ .792,  0,  0}{\textbf{0.871}} \\
			& ${F_\beta ^\omega}$ ~\hfill\hfill$\uparrow$   & 0.755  & \textcolor[rgb]{ 0,  .533,  .2}{\textbf{0.804}} & 0.770  & 0.771  & \textcolor[rgb]{ 0,  .4,  .8}{\textbf{0.803}} & 0.784  & \textcolor[rgb]{ .792,  0,  0}{\textbf{0.814}} \\
			& $MAE$\hfill$\downarrow$  & 0.050  & \textcolor[rgb]{ 0,  .533,  .2}{\textbf{0.043}} & 0.049  & 0.045  & \textcolor[rgb]{ 0,  .4,  .8}{\textbf{0.044}} & \textcolor[rgb]{ 0,  .533,  .2}{\textbf{0.043}} & \textcolor[rgb]{ .792,  0,  0}{\textbf{0.041}} \bigstrut[b]\\
			\hline
			\multirow{4}[2]{*}{\begin{sideways}HKU-IS\end{sideways}} & ${E_\xi}$ ~\hfill\hfill$\uparrow$   & 0.953  & \textcolor[rgb]{ 0,  .533,  .2}{\textbf{0.965}} & 0.956  & \textcolor[rgb]{ 0,  .4,  .8}{\textbf{0.960}} & \textcolor[rgb]{ 0,  .533,  .2}{\textbf{0.965}} & 0.959  & \textcolor[rgb]{ .792,  0,  0}{\textbf{0.969}} \bigstrut[t]\\
			& ${S_\alpha}$ ~\hfill\hfill$\uparrow$   & 0.928  & \textcolor[rgb]{ 0,  .533,  .2}{\textbf{0.935}} & 0.924  & 0.927  & \textcolor[rgb]{ 0,  .4,  .8}{\textbf{0.932}} & 0.931  & \textcolor[rgb]{ .792,  0,  0}{\textbf{0.939}} \\
			& ${F_\beta ^\omega}$ ~\hfill\hfill$\uparrow$   & 0.897  & \textcolor[rgb]{ 0,  .4,  .8}{\textbf{0.925}} & 0.908  & 0.917  & \textcolor[rgb]{ 0,  .533,  .2}{\textbf{0.932}} & 0.915  & \textcolor[rgb]{ .792,  0,  0}{\textbf{0.938}} \\
			& $MAE$\hfill$\downarrow$  & 0.029  & \textcolor[rgb]{ 0,  .4,  .8}{\textbf{0.022}} & 0.026  & 0.023  & \textcolor[rgb]{ 0,  .533,  .2}{\textbf{0.020}} & 0.024  & \textcolor[rgb]{ .792,  0,  0}{\textbf{0.018}} \bigstrut[b]\\
			\hline
			\multirow{4}[2]{*}{\begin{sideways}PASCAL-S\end{sideways}} & ${E_\xi}$ ~\hfill\hfill$\uparrow$   & 0.843  & \textcolor[rgb]{ 0,  .4,  .8}{\textbf{0.875}} & 0.870  & 0.870  & 0.873  & \textcolor[rgb]{ 0,  .533,  .2}{\textbf{0.879}} & \textcolor[rgb]{ .792,  0,  0}{\textbf{0.880}} \bigstrut[t]\\
			& ${S_\alpha}$ ~\hfill\hfill$\uparrow$   & 0.871  & \textcolor[rgb]{ .792,  0,  0}{\textbf{0.885}} & 0.864  & 0.871  & 0.878  & \textcolor[rgb]{ 0,  .4,  .8}{\textbf{0.881}} & \textcolor[rgb]{ 0,  .533,  .2}{\textbf{0.883}} \\
			& ${F_\beta ^\omega}$ ~\hfill\hfill$\uparrow$   & 0.822  & \textcolor[rgb]{ 0,  .4,  .8}{\textbf{0.860}} & 0.833  & 0.844  & \textcolor[rgb]{ 0,  .533,  .2}{\textbf{0.862}} & 0.854  & \textcolor[rgb]{ .792,  0,  0}{\textbf{0.866}} \\
			& $MAE$\hfill$\downarrow$  & 0.061  & \textcolor[rgb]{ 0,  .533,  .2}{\textbf{0.048}} & 0.061  & 0.054  & \textcolor[rgb]{ 0,  .4,  .8}{\textbf{0.049}} & 0.051  & \textcolor[rgb]{ .792,  0,  0}{\textbf{0.047}} \bigstrut[b]\\
			\hline
		\end{tabular}%
	}
	\label{tab:compare_rgb}%
\end{table}%
In order to demonstrate the advantages and applications of our method in more scenarios, we also compare it with some recent advanced single-modal SOD methods, with the results reported in Table~\ref{tab:compare_rgb}. Considering that the single-modal SOD task only uses the RGB modality, we replace the original RGBT image pair with two identical RGB images as input to our network. 

\textit{1) Datasets:}
We evaluate our method on five representative single-modal SOD datasets, including DUTS~\cite{wang2017Learning} (10,553 training images and 5,019 testing images), OMRON~\cite{yang2013saliency} (5,168 images), ECSSD~\cite{yan2013Hierarchical} (1,000 images), HKU-IS~\cite{li2015visual} (4,447 images), and PASCAL-S~\cite{li2014Secrets} (850 images). Following~\cite{zhuge2022salient,yun2023towards}, we use the training set of DUTS to train our single-modal SOD model.

\textit{2) Experiment Setup:}
We compare our method with 6 state-of-the-art single-modal SOD methods, including VST~\cite{liu2021visual}, ICON~\cite{zhuge2022salient}, END~\cite{wu2022edn}, MENet~\cite{wang2023pixels}, BBRF~\cite{ma2023bbrf}, and SelfReformer~\cite{yun2023towards}. All predicted results and code used in the experiments are released by the authors.

\textit{3) Quantitative Evaluation:}
The results in Table~\ref{tab:compare_rgb} show that our method outperforms all compared methods on the five datasets, except for the S-measure metric on the PASCAL-S dataset. For example, compared with the second best method (i.e., ICON), our method achieves an average improvement of 0.5\%, 0.3\%, 1.3\%, and 11.8\% for the four evaluation metrics (i.e., ${E_\xi }$, ${S_\alpha }$, ${F_\beta ^\omega}$, and $MAE$) across the five datasets. This demonstrates that our method is also applicable to the single-modal SOD task.

\subsection{Failure Cases and Future Work}
\begin{figure}[t]
	\centering
	\includegraphics[width=1\columnwidth]{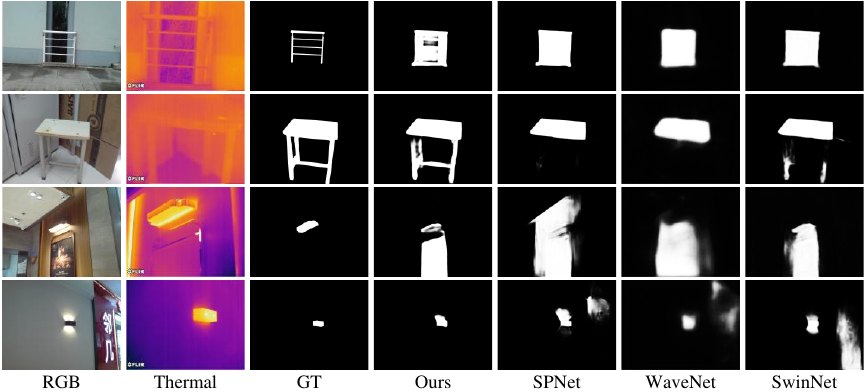}
	\caption{Visual results of our SACNet and other advanced methods (i.e., SPNet~\cite{zhang23SPNet}, WaveNet~\cite{zhou2023wave}, and SwinNet~\cite{liu2021swinnet}) in some typical failure cases, including hollow objects (i.e., Rows 1 and 2), strong light interference (i.e., Rows 3 and 4).}
	\label{fig::failure}
\end{figure}
Although our method achieves superior performance on unaligned image pairs through modeling robust multi-modal correlations, it still fails in some extreme challenging scenarios. Fig.~\ref{fig::failure} shows the visual results of our method and other advanced methods (i.e., SPNet~\cite{zhang23SPNet}, WaveNet~\cite{zhou2023wave}, and SwinNet~\cite{liu2021swinnet}) for some typical failure cases. For the examples in the first and second rows, the salient objects are hollow and their foreground and background regions are intertwined, which makes it difficult for our method and other methods to make refined predictions. The salient objects in the third and forth rows are challenged by strong light, which interferes with our correlation modeling of salient regions, resulting in inaccurate localization and segmentation of the salient objects. 
These failure cases reveal the limitations of our method, which still struggles to deal with the challenging scenarios in unaligned image pairs. To this end, in future work, we will design fusion modules to handle different challenges. By integrating the feature representations of these specific fusion modules into the correlation modeling, the robustness of our method in challenging scenarios can be further improved.

In addition, UVT2000 serves as the first unaligned RGBT SOD dataset for research on alignment-free RGBT SOD, it still has some limitations. We will further expand the UVT2000 dataset in future work to better reflect the diversity and complexity in real-world applications. Specifically, in terms of scale, we will capture larger-scale unaligned RGBT image pairs that far exceed the size of any existing multi-modal SOD datasets to improve the diversity of the UVT2000 dataset. In terms of scenes and circumstances, we will cover more practical scenes, such as traffic scenes, drone scenes, and lake scenes, to promote extensive research on unaligned RGBT image pairs for real-world applications. In terms of challenges, we will capture and annotate more representative challenging scenarios that are specific to unaligned RGBT image pairs to enhance the complexity of the UVT2000 dataset. In this way, the UVT2000 dataset will be improved as a more solid foundation for comprehensive research on alignment-free RGBT SOD.

\section{Conclusion}
In this paper, we explore the saliency complementarity in unaligned visible-thermal image pairs and propose a semantics-guided asymmetric correlation network, which models robust correlations between RGB and thermal modalities without manual alignment. To this end, two components (i.e., ACM and AFSM) are proposed. The ACM is able to establish comprehensive multi-modal correlations specific to salient regions, and the AFSM can sample relevant thermal features conditioned on corresponding RGB features for accurate integration. Additionally, we contribute a novel unaligned RGBT SOD benchmark dataset called UVT2000, which provides a challenging platform and facilitates research on alignment-free RGBT SOD. Experimental results demonstrate that our method achieves state-of-the-art performance on both aligned and unaligned datasets. The overall performance of existing methods and our method on the newly constructed UVT2000 dataset shows the great potential of alignment-free RGBT SOD.

\bibliographystyle{IEEEtran}
\bibliography{mybibfile}

\end{document}